\documentclass[twoside]{article}

\usepackage[accepted]{aistats2024}
\usepackage[round]{natbib}

\makeatletter
\newcommand{\maketitlenew}{\@maketitle}
\makeatother

\usepackage{algorithm}
\usepackage{algpseudocode}

\usepackage{chemarr} 
\usepackage{amssymb}
\usepackage{amsmath}
\usepackage{amsfonts}       
\usepackage{mathtools}
\usepackage{nicefrac}       
\usepackage{amsthm}         
\usepackage{dsfont}
\usepackage{empheq}

\allowdisplaybreaks

\theoremstyle{plain}

\theoremstyle{definition}

\theoremstyle{remark}

\usepackage{graphicx}
\graphicspath{ {figures/} }
\usepackage{epsfig}
\usepackage{subcaption}
\usepackage{float}
\usepackage{wrapfig}

\usepackage[utf8]{inputenc} 
\usepackage[T1]{fontenc}    
\usepackage{microtype}      
\usepackage{textcomp}

\usepackage{booktabs}       
\usepackage{makecell}

\usepackage{color}
\usepackage{xcolor}
\usepackage{xkeyval}

\usepackage{tikz}
\usetikzlibrary{bayesnet}
\usetikzlibrary{positioning}
\usetikzlibrary{arrows}
\usetikzlibrary{calc}

\usetikzlibrary{positioning,fit,shapes,backgrounds,circuits.logic.US}

\tikzstyle{server}=[circle, line width=0.5pt, rounded corners=0.1mm, draw=black!100, fill=tud3a!100]
\tikzstyle{vertex}=[circle, line width=1.5pt, draw=tud0d, fill=white]
\tikzstyle{dispatcher} =[and gate US, line width=0.5pt, draw=black!100, fill=tud0c!100]
\tikzstyle{dotbox} = [draw=white, fill=white, rectangle,  inner sep=10pt, inner ysep=20pt]

\tikzset{three_sided/.style={
		draw=none,rectangle, 
		append after command={
			[shorten <= -0.5\pgflinewidth]
			([shift={(-1.5\pgflinewidth,-0.5\pgflinewidth)}]\tikzlastnode.north west)
			edge([shift={( 0.5\pgflinewidth,-0.5\pgflinewidth)}]\tikzlastnode.north east)
			([shift={( 0.5\pgflinewidth,-0.5\pgflinewidth)}]\tikzlastnode.north east)
			edge([shift={( 0.5\pgflinewidth,+0.5\pgflinewidth)}]\tikzlastnode.south east)
			([shift={( 0.5\pgflinewidth,+0.5\pgflinewidth)}]\tikzlastnode.south east)
			edge([shift={(-1.0\pgflinewidth,+0.5\pgflinewidth)}]\tikzlastnode.south west)
		}
	}
}

\usepackage{paralist} 
\usepackage{amstext} 
\usepackage{array} 
\usepackage{hyperref}       

\usepackage{url}            
\usepackage{xspace}
\usepackage{csquotes}
\usepackage{etoolbox}
\usepackage[shortlabels,inline]{enumitem}  
\newlist{inlineitemize}{enumerate*}{1}
\setlist[inlineitemize]{label=(\roman*)}
\usepackage{calc}

\usepackage[capitalize, nameinlink]{cleveref}
\crefname{appendix}{Appendix}{Appendecies}

\usepackage{ifthen}
\usepackage{balance}

\usepackage{todonotes}
\newboolean{todo}
\setboolean{todo}{true} 

\newcommand{\remarkInternal}[4]{\ifthenelse{\boolean{todo}}{\todo[inline, color=#2, caption={2do}, #3]{\begin{minipage}{\textwidth-4pt}\emph{Remark #1:}\\#4\end{minipage}}}{}}

\usepackage{blindtext}
\newboolean{blind}
\setboolean{blind}{false} 
\newcommand{\blindparagraph}[2][]{\ifthenelse{\boolean{blind}}{\blindtext[1]}{}}

\usepackage{hyphenat} 
\newcommand{\eg}{e.g.\@\xspace}
\newcommand{\ie}{i.e.\@\xspace}
\newcommand{\wrt}{w.r.t.\@\xspace}
\newcommand{\rhs}{r.h.s.\@\xspace}

\DeclareMathOperator{\diag}{diag}

\DeclareMathOperator*{\argmin}{\arg\min}
\DeclareMathOperator*{\argmax}{\arg\max}

\DeclareMathOperator{\1}{\mathds{1} }

\DeclareMathOperator{\E}{\mathsf{E}}

\newcommand{\Eof}{\E\expectarg}
\DeclarePairedDelimiterX{\expectarg}[1]{[}{]}{%
    \ifnum\currentgrouptype=16 \else\begingroup\fi
    \activatebar#1
    \ifnum\currentgrouptype=16 \else\endgroup\fi
}
\newcommand{\innermid}{\nonscript\;\delimsize\vert\nonscript\;}
\newcommand{\activatebar}{%
    \begingroup\lccode`\~=`\|
    \lowercase{\endgroup\let~}\innermid
    \mathcode`|=\string"8000
}

\DeclarePairedDelimiterX{\expectargleft}[1]{[}{.}{%
    \ifnum\currentgrouptype=16 \else\begingroup\fi
    \activatebar#1
    \ifnum\currentgrouptype=16 \else\endgroup\fi
}

\DeclarePairedDelimiterX{\expectargright}[1]{.}{]}{%
    \ifnum\currentgrouptype=16 \else\begingroup\fi
    \activatebar#1
    \ifnum\currentgrouptype=16 \else\endgroup\fi
}

\DeclareMathOperator{\KL}{\mathsf{KL}}

\newcommand{\KLof}{\KL\klarg}
\DeclarePairedDelimiterX{\klarg}[1]{(}{)}{%
    \ifnum\currentgrouptype=16 \else\begingroup\fi
    \activatediv#1
    \ifnum\currentgrouptype=16 \else\endgroup\fi
}
\newcommand{\innerdiv}{\nonscript\;\delimsize\Vert\nonscript\;}
\newcommand{\activatediv}{%
    \begingroup\lccode`\~=`\|
    \lowercase{\endgroup\let~}\innerdiv
    \mathcode`|=\string"8000
}

\DeclarePairedDelimiterX{\klargleft}[1]{()}{.}{%
    \ifnum\currentgrouptype=16 \else\begingroup\fi
    \activatediv#1
    \ifnum\currentgrouptype=16 \else\endgroup\fi
}

\DeclarePairedDelimiterX{\klargright}[1]{.}{)}{%
    \ifnum\currentgrouptype=16 \else\begingroup\fi
    \activatediv#1
    \ifnum\currentgrouptype=16 \else\endgroup\fi
}

\DeclareMathOperator{\MultDis}{Mult}
\DeclareMathOperator{\BinDis}{Bin}

\DeclareMathOperator{\PoisDis}{Pois}

\newcommand{\prob}{\ensuremath{\mathit{p}}}
\newcommand{\qrob}{\ensuremath{\mathit{q}}}

\DeclareMathOperator{\Prob}{P}

\newcommand{\diff}{\mathop{}\!\mathrm{d}}

\usepackage[nolist]{acronym}

\begin{document}

\begin{acronym}
    \acro{mjp}[MJP]{Markov jump process}
    \acroplural{mjp}[MJPs]{Markov jump processes}
    \acroindefinite{mjp}{an}{a}
    \acro{lqg}[LQG]{linear quadratic Gaussian}
    \acroindefinite{lqg}{an}{a}
    \acro{pomdp}[POMDP]{partially observable Markov decision process}
    \acroplural{pomdp}[POMDPs]{partially observable Markov decision processes}
    \acro{hjb}[HJB]{Hamilton-Jacobi-Bellman}
    \acroindefinite{hjb}{an}{a}
    \acro{pde}[PDE]{partial differential equation}
    \acro{ctmc}[CTMC]{continuous-time Markov chain}
    \acroplural{ctmc}[CTMCs]{continuous-time Markov chains}
    \acro{mdp}[MDP]{Markov decision process}
    \acroplural{mdp}[MDPs]{Markov decision processes}
    \acroindefinite{mdp}{an}{a}
    \acro{smdp}[SMDP]{semi-Markov decision process}
    \acroplural{smdp}[SMDPs]{semi-Markov decision processes}
    \acroindefinite{smdp}{an}{a}
    \acro{ode}[ODE]{ordinary differential equation}
    \acroindefinite{ode}{an}{an}
    \acro{kl}[KL]{Kullback-Leibler}
     \acro{pomcp}[POMCP]{partially observable Monte Carlo planning}
      \acro{despot}[DESPOT]{determinized sparse partially observable Tree}
      \acro{crn}[CRN]{chemical reaction network}
      \acroplural{crn}[CRNs]{chemical reaction networks}
      \acro{LV}[LV]{Lotka-Volterra}
\end{acronym}
\twocolumn[

\aistatstitle{Approximate Control for Continuous-Time POMDPs}

\aistatsauthor{{Yannick Eich}\And{Bastian Alt}\And{Heinz Koeppl}}

\aistatsaddress{Department of Electrical Engineering and Information Technology\\\
  Technische Universität Darmstadt
  \\
  \texttt{\{yannick.eich, bastian.alt, heinz.koeppl\}@tu-darmstadt.de} }

  
   ]

\begin{abstract}
This work proposes a decision-making framework for partially observable systems in continuous time with discrete state and action spaces. As optimal decision-making becomes intractable for large state spaces we employ approximation methods for the filtering and the control problem that scale well with an increasing number of states. Specifically, we approximate the high-dimensional filtering distribution by projecting it onto a parametric family of distributions, and integrate it into a control heuristic based on the fully observable system to obtain a scalable policy. We demonstrate the effectiveness of our approach on several partially observed systems, including queueing systems and chemical reaction networks.
\end{abstract}
\section{\uppercase{Introduction}}
Partial observability in dynamical systems is a ubiquitous problem for many applications.
This includes settings such as robotics \citep{lauri2022partially}, communication systems and signal processing \citep{proakis2008digital,kay1993fundamentals}, or biology \citep{wilkinson2018stochastic}.
In partially observable settings only noisy data from a latent time-dependent process is available.
A principled way to deal with the resulting inference problem is Bayesian filtering \citep{bain2009fundamentals,sarkka2013bayesian}.
The framework of Bayesian filtering can be exploited to infer in an online manner the latent state of the system given the historical information available. 
The information of the latent state is then encoded in the filtering posterior distribution, the belief state.
This stochastic filtering approach is especially appealing for the control of such partially observed dynamical systems.
This includes among others, \eg, control problems with noisy sensor measurements, such as grasping and navigation in robotics \citep{kurniawati2008sarsop} or cognitive medium access control \citep{zhao2005decentralized} for communication systems.
For finding decision strategies, which use the available observational data to control the system at hand, a solid framework can be found in the area of optimal control \citep{stengel1994optimal}.
The classical setting for partially observable problems in optimal control theory is historically a continuous-time setting with a real-valued stochastically evolving latent state \citep{bensoussan1992stochastic}. 
This includes the well-known \ac{lqg} control problem, dating back to the work of \citet{wonham1968separation}.
Contrary to the continuous modeling approach, a discrete-time and discrete-state setting has historically been discussed in the area of operations research \citep{aastrom1965optimal}, and has ever since found popularity as \ac{pomdp} within the machine learning community \citep{kaelbling1998planning}. 

Over the years numerous algorithms for solving partially observable control problems, especially for discrete-time settings have been proposed.
For example, the work of \citet{zhou2010solving}, which is closely related to ours, uses projection filtering to represent the belief to perform optimal control for continuous state and action spaces.
Similarly, other approximate filtering methods, such as particle filtering \citep{thrun1999monte}, have been exploited in a discrete-time context.
For the discrete-time setting with discrete spaces, Monte Carlo tree search methods have shown substantial success, such as \acs{despot} \citep{somani2013despot} and \acs{pomcp} \citep{silver2010monte}.
Also, more recently methods solving \acp{pomdp} in a discrete-time regime using modern deep-learning techniques such as the algorithms of \citet{igl2018deep} and \citet{singh2021structured}, have shown to find good control strategies.
However, many problems can neither be modeled by a discrete-time approach nor using continuous-time and space strategies such as the \ac{lqg} setup.
Consider for example the control of a queueing network within a communication system \citep{bolch2006queueing}. 
Here the state space is discrete as the system is described by the discrete number of packets in each queue.
However, as an instance of a discrete event system, it can not be directly described by a discrete-time modeling approach as the packets arrive and are serviced at non-equidistant time points.
Another example is the modeling of molecules in a biological system. 
In the context of systems biology \citep{wilkinson2018stochastic} the dynamical system is described by the discrete number of each of the molecules in the system.
The molecule count is dynamically changed due to reactions occurring in the system, which as a physical system evolves continuously in time.
When the control of such systems is of interest, it poses a problem for both classical setups and the derived algorithms.
Even though there exist strategies to transform these continuous-time discrete-state space problems to one of both setups they are often not sensible and subject to large modeling errors.
For example, Langevin dynamics or a deterministic fluid limit are methods used to approximate discrete states by continuous ones, see, \eg, \citet{gardiner2009stochastic} and \citet{ethier2009markov}.
Although these approximations have been used, \eg, in the context of heavy traffic in queueing networks \citep{kushner2001heavy}, they fail when the latent state is inherently discrete.
This is the case, when the ordinal number representing the state is small or in the extreme case when on-off switching behavior occurs in the system, as the state can not sufficiently be described by a continuous value.
Naively discretizing time also has its downside, as many algorithms derived under a discrete-time assumptions are often not robust to time discretization \citep{tallec2019making}.
Contrary to that, continuous-time models can numerically be solved using elaborate numerical differential equation solvers, that automatically adapt, \eg, the step size to the specific problem.

The control for fully observed continuous-time discrete-state space \acp{mdp} has been discussed in the context of \acp{smdp}, \eg, for the control of queueing networks \citep{bertsekas2012adynamic,bertsekas2012bdynamic}, see also \citet{du2020model} and the references therein.
However, partial observability in this setting has received little attention by the machine learning community.
Recently, \citet{alt2020pomdps} discussed the theory to model partially observed continuous-time discrete-state space problems. 
However, a substantial scalability issue of their approach is that it has to solve a high-dimensional \ac{pde} in belief space.
Additionally, the presented approach leverages exact filtering within the control problem. 
These two algorithmic choices make the presented method doubly intractable, as both the exact filter and the control problem in belief space suffer from the curse of dimensionality. 

Hence, our contributions are:
We provide a new scalable algorithm for the solution of continuous-time discrete-state space \acp{pomdp}. 
Our new method can be divided into two parts:
First, we approximate the filtering distribution by a parametric distribution using the method of entropic matching \citep{bronstein2018variational}.
Here we give closed-form solutions for the evolution of the parameters in some interesting problem settings, such as queueing networks and \acp{crn}.
Second, to get a scalable control law we adapt a heuristic used historically within discrete-time partially observable systems to the continuous-time \ac{pomdp} framework described in \citet{alt2020pomdps}.
This enables us to consider control problems with an unbounded number of states, well beyond the setup of \citet{alt2020pomdps}. An implementation of our proposed method is publicly available\footnote{\url{https://github.com/yannickeich/ApproxPOMDPs}}.
\section{\uppercase{Model}}\label{sec:model}
We consider the problem of optimal decision-making under partial observability in continuous time $t \in \mathbb R_{\geq 0}$.
For this, we exploit a continuous-time \ac{pomdp} model \citep{alt2020pomdps}, where the latent state trajectory $X_{[0,\infty)} \coloneqq \{X(t) \in \mathcal X \mid t \in \mathbb R_{\geq 0}\}$ is modeled as a controlled \ac{ctmc} \citep{norris1998markov} on a countable state space $\mathcal X \subseteq \mathbb N^n$.
The rate function for a state $x \neq x'$ of the \ac{ctmc} is given as
\begin{equation*}
\begin{multlined}
        \Lambda(x,x', u, t) \\ 
        \coloneqq \lim_{h \to 0} h^{-1} \Prob(X(t+h)=x' \mid X(t)=x, u(t)=u),
\end{multlined}
\end{equation*}
where we assume time homogeneity, \ie, $\Lambda(x,x', u, t) \equiv \Lambda(x,x', u), \forall t$.
In the above equation the state trajectory $X_{[0,\infty)}$ can be controlled by a control trajectory $u_{[0,\infty)} \coloneqq \{u(t) \in \mathcal U \mid t \in \mathbb R_{\geq 0}\}$. 
Throughout this work, we assume a finite action space setting, \ie, $\mathcal U=\{1, 2, \dots, m\}$, where $m$ denotes the number of actions.
We assume that $X(t)$ can not be directly observed but only a partial observation $Y(t) \in \mathcal Y$ is available.
The goal of an optimal control sequence is then to maximize a reward function $R:\, \mathcal X \times \mathcal U \rightarrow \mathbb R$
over a time horizon, \ie,
\begin{equation*}
    \begin{aligned}
    &\underset{u_{[0,\infty)}}{\text{maximize}} &&\Eof*{\int_0^\infty e^{-\frac{t}{\tau}} R(X(t),u(t))\diff t},
    \end{aligned}
\end{equation*}
where we assume an infinite horizon setting with inverse exponential discount variable $\tau \in \mathbb R_{\geq 0 }$.
Note that, in the \ac{pomdp} setting the control $u(t)$ at time point $t$ can only depend on the observation and control history $y_{[0,t]} \coloneqq \{y(s) \mid s \in [0,t] \}$ and $u_{[0,t)}\coloneqq \{u(s) \mid s \in [0,t) \}$, respectively, \ie, $u_{[0,\infty)}$ has to be an admissible control trajectory.
Mathematically, we let $u_{[0,\infty)} \in \mathcal D_{\mathcal F_y^u}([0,\infty), \mathcal U)$, where $\mathcal D_{\mathcal F_y^u}([0,\infty), \mathcal U)$ is the space of càdlàg functions on $[0,\infty)$ taking values in $\mathcal U$ that are adapted to the family of control dependent sigma-algebras $\mathcal F_y^u(t)=\sigma(Y(s)\mid s \leq t)$ of the observation process, \ie, indicating that $u(t)$ is $\mathcal F_y^u(t)$-measurable \citep{bensoussan1992stochastic}.

It is well known that the partial observable control problem can be cast as a problem of optimal control in belief space \citep{bertsekas2012adynamic,bertsekas2012bdynamic}.
The belief at time point $t$ is the Bayesian filtering distribution
\begin{equation*}
    \pi_t(x)=\Prob(X(t)=x \mid y_{[0,t]}, u_{[0,t)}).
\end{equation*}
This yields the optimal feedback policy $u(t)=\mu^\ast(\pi_t)$.

In the next sections, we explain how to find both the optimal feedback policy, as well as how to compute the exact filtering distribution. 
However, both problems are computationally intractable.
Hence, we present two approximations for the solution of 
\begin{inlineitemize}
\item the optimal filtering problem and
\item the optimal control problem.
\end{inlineitemize}

\section{\uppercase{Continuous-Time Probabilistic Inference for POMDPs}}
The state of the underlying belief \ac{mdp} is characterized by the filtering distribution $\pi_t(x)$.
For computing the filtering distribution, we first consider the prior distribution for the latent process $X_{[0,\infty)}$.
We can characterize it by its time-point-wise marginal distribution $\prob_t(x) \coloneqq \Prob(X(t)=x \mid  u_{[0,t)})$.
For a controlled \ac{ctmc} $X_{[0,\infty)}$, with action trajectory $u_{[0,\infty)}$ and rate function $\Lambda$, the time evolution of the prior is given by the differential form of the forward Chapman-Kolmogorov equation \citep{ethier2009markov}, the \emph{master equation} \citep{gardiner2009stochastic},
\begin{equation}
\begin{aligned}
     &\frac{\diff}{\diff t} \prob_t(x) = [\mathcal L_{u(t)} \prob_t](x)\
\end{aligned}
\label{eq:master_eq}
\end{equation}
with initial distribution $\prob_0(x)$ at time point $t=0$.
The evolution operator $\mathcal L_{u}$ of the Markov chain is given as
\begin{equation*}
\begin{multlined}
    [\mathcal L_{u} \phi](x) \\
    \coloneqq \sum_{x' \neq x}\big \{ \Lambda(x',x,u) \phi(x')- \Lambda(x,x',u) \phi(x)\big\}
\end{multlined}
\end{equation*}
for an arbitrary test function $\phi$.
Next, we have to specify the generative model for the observation process $\{Y(t) \mid t \in \mathbb R_{\geq 0}\}$.
In this paper, we will consider two observation models, which are
\begin{inlineitemize}
    \item a discrete-time noisy measurement model and
    \item a sub-system measurement model.
\end{inlineitemize}

\paragraph{Noisy Measurements.}
A natural observation model to consider is a discrete-time noisy measurement model, which we denote as $D$.
For this we assume that the observations $Y(t)$ are given at discrete time instances $\{t_i\}$, \ie, $\{Y_i \coloneqq Y(t_i) \mid i \in \mathbb N \}$.
Further, we assume that the state is not directly observed but only a noisy measurement is available as
\begin{equation}
\begin{aligned}
    &D:  &&Y_i \mid \{ X(t_i)=x, u(t_i^-)=u\} \sim \prob(y_i \mid x, u),
\end{aligned}
\label{eq:model_d}
\end{equation}
where throughout this paper we denote by $\phi(t_i^-) \coloneqq \lim_{t \nearrow t_i} \phi(t)$ the limit from the left of an arbitrary function $\phi$, respectively.
Note that this observation model gives rise to a continuous-discrete filtering problem, \ie, filtering for latent states evolving in continuous-time and discrete-time observations, for more see, \eg, \citet{maybeck1982stochastic}, and \citet{sarkka2019applied}. 
\paragraph{Sub-System Measurements.}
Additionally we consider a sub-system measurement model, which we denote by $C$.
For this we assume that we can only observe some components of the $n$-dimensional random vector $X(t)$. 
Consider without loss of generality that $X(t)=[\hat X^\top (t), \bar X^\top(t)]^\top$, and that the components $\hat X(t)$ and $\bar X(t)$ are unobserved and observed, respectively. 
Hence, as for the observations $\{Y(t) \mid t \in \mathbb R_{\geq 0} \}$ we have the noise-free measurement model 
\begin{equation}
\begin{aligned}
    &C:  &&Y(t)=\bar X(t).
\end{aligned}
\label{eq:model_c}
\end{equation}
As the process $\{X(t)\}$ evolves on a countable space, this implies that at the discrete-time instances $\{t_i\}$ we observe a jump from a state $\bar X(t_i^-)$ to state $\bar X(t_i)$, \ie, we have a discrete-time noise-free observation as $Y_i \coloneqq Y(t_i)=\bar X(t_i)$.
Note that this observation model \citep{bronstein2018marginal} can be seen as a generalization of a Poisson process observation model \citep{elliott2005general}.

\subsection{Exact Inference}\label{sec:exact_inf}
When considering the task of computing the filtering distribution, the sought after posterior distribution can be calculated by Bayes' rule.
This yields for the noisy measurement model $D$ in \cref{eq:model_d} a differential equation for the filter between observations as
\begin{equation}
\begin{aligned}
    &D: &&\frac{\diff}{\diff t} \pi_t(x) = [\mathcal L_{u(t)} \pi_t](x).
\end{aligned}
\label{eq:pred_d}
\end{equation}
At observation time points, Bayes' rule is computed as
\begin{equation}
\begin{aligned}
      &D: &&\pi_{t_i}(x) = Z_i^{-1} \prob(y_i \mid x , u(t_i^-)) \pi_{t_i^-}(x),
\end{aligned}
\label{eq:update_d}
\end{equation}
where $Z_i=\sum_{x} \prob(y_i \mid x , u(t_i^-)) \pi_{t_i^-}(x)$ is a normalization constant, for more see \citet{huang2016reconstructing}.
The resulting filtering equation is the usual result in continuous-discrete filtering, as the \emph{prediction step} in \cref{eq:pred_d} between the observations is given by the time evolution of the prior distribution, see \cref{eq:master_eq}, and the \emph{update step} in \cref{eq:update_d} carries out Bayes' rule for the observation $y_i$.

For the sub-system measurement model $C$ in \cref{eq:model_c}, we find analog equations between jumps of the observation process, \ie, when the observation process is constant $y(t)=y(t^-)$,
\begin{equation}
\begin{aligned}
    &C: &&\begin{multlined}     
        \frac{\diff }{\diff t} \pi_t(x) = \1_{y(t)}(x) [\mathcal L_{u(t)} \pi_t](x) \\
    -\pi_t(x)\sum_{x'} \1_{y(t)}(x') [\mathcal L_{u(t)} \pi_t](x'),
    \end{multlined}
\end{aligned}
\label{eq:pred_c}
\end{equation}
where $x=[\hat x^\top, \bar x^\top]^\top$ and $\1_{y(t)}(x) \coloneqq \1(y(t)=\bar x)$.
When the observation process jumps $y_i=y(t_i) \neq y(t_i^-)$, we have
\begin{equation}
\begin{aligned}
      &C: &&\pi_{t_i}(x)=Z_i^{-1} \1_{y_i}(x) [\mathcal L_{u(t)} \pi_{t_i^-}](x),
\end{aligned}
\label{eq:update_c}
\end{equation}
with the normalization constant $Z_i=\sum_x \1_{y_i}(x) [\mathcal L_{u(t)} \pi_{t_i^-}](x)$, for more see \citet{bronstein2018marginal}.

However, all these equations suffer from the curse of dimensionality as the numerical solution of the system of \acp{ode} in \cref{eq:pred_d,eq:pred_c} scales in $\vert \mathcal X \vert^2$, which in the case $\mathcal X = \mathbb N^n$ is even infinite. 
Similarly, computing the update steps in \cref{eq:update_d,eq:update_c} requires computing the normalization constant $Z_i$, which is intractable.
Hence, we propose next an approximate inference method for the solution to the intractable exact filtering problem.

\subsection{Approximate Inference}\label{sec:approx_inf}
In order to overcome the challenge of computing the exact intractable filtering distribution, we propose to approximate it by a parametric distribution using an assumed density filtering method \citep{maybeck1982stochastic}.
For this we use a deterministic approximation method, entropic matching \citep{PhysRevE.87.022719,bronstein2018variational}, which can be seen as a continuous-time extension of a special expectation propagation method \citep{minka2005divergence}.
As a result, we obtain low-dimensional time evolution equations for the parameters of the distribution. 

We start by assuming a parametric form for the filtering distribution at time $t$ as
    $\pi_t(x) \approx \qrob_{\theta(t)}(x)$,
with parameters $\theta(t) \in \Theta \subseteq \mathbb R^{d_{\theta}}$. We then consider the evolution of the distribution $\qrob_{\theta(t)}(x)$ over a small time step $h$ in absence of new measurements.
For the observation model $D$, the distribution follows the prediction step in \cref{eq:pred_d}, which yields
\begin{equation*}
    \tilde\pi_{t+h}(x) \coloneqq \qrob_{\theta(t)}(x) + h [\mathcal L_{u(t)} \qrob_{\theta(t)}](x) + o(h).
\end{equation*}
It is worth noting that $\tilde\pi_{t+h}(x)$ generally does not match the structure of the parametric family used for approximation.
To obtain a distribution within the parametric family, its optimal parameters after a small time step $h$ can be computed by minimizing the reverse \ac{kl} divergence 
\begin{equation*}
    \theta(t+h)=\argmin_{\theta'} \KLof*{\tilde\pi_{t+h} | \qrob_{\theta'}}.
\end{equation*}
The entropic matching method approximates the change of the filtering distribution in an infinitesimal time step such that the approximated distribution stays in the space of the parametric distributions.
By computing the continuous-time limit $\lim_{h \to 0} h^{-1}\{\theta(t+h)-\theta(t)\}$ \iac{ode} for the parameters can be found \citep{bronstein2018variational} as
\begin{equation}
\begin{aligned}
    &D: &&\begin{multlined}\frac{\diff }{\diff t} \theta(t) = F(\theta(t))^{-1}\\
     \cdot \E_{\qrob} \left[\mathcal L_{u(t)}^\dagger \nabla_{\theta(t)} \log \qrob_{\theta(t)}(X(t))\right],
    \end{multlined}
\end{aligned}
\label{eq:em_pred_d}
\end{equation}
where $F(\theta)$ is the Fisher information matrix of the parametric distribution, \ie, $F(\theta)= \E_{q}\left[\nabla_\theta \log \qrob_{\theta}(X) \nabla_\theta^\top \log \qrob_{\theta}(X)\right]$. The score $\nabla_{\theta(t)} \log \qrob_{\theta(t)}(x)$ evolves according to the operator $\mathcal L_{u}^{\dagger}$
\begin{equation*}
   [\mathcal L_{u}^{\dagger} \psi](x) \coloneqq \sum_{x' \neq x} \Lambda(x, x', u)\left\{\psi(x') - \psi(x)\right\},
\end{equation*}
for an arbitrary test function $\psi$, where $\mathcal L_{u}^{\dagger}$ is adjoint to the evolution operator $\mathcal L_{u}$, \wrt the inner product space $\langle \phi, \psi \rangle=\sum_x \phi(x) \psi(x)$.
At the observation jump time points, we have a discontinuity in the filtering distribution as in the update step in \cref{eq:update_d}. Here we find the optimal parameters accordingly as 
\begin{equation}
    \theta(t_i)= \argmin_{\theta'} \KLof*{\hat\pi_{t_i} | \qrob_{\theta'}},
    \label{eq:em_update}
\end{equation}
where $\hat\pi_{t_i}$ is the posterior distribution that follows from Bayes' rule
\begin{equation*}
\begin{aligned}
    &D: &&\hat\pi_{t_i}(x) \coloneqq Z_i^{-1}  \prob(y_i \mid x , u(t_i^-)) \qrob_{\theta(t_i^-)}(x),
\end{aligned}
\label{eq:em_update_d}
\end{equation*}
with $Z_i = \sum_x \prob(y_i \mid x , u(t_i^-)) \qrob_{\theta(t_i^-)}(x)$.
For the observation model $C$, we have for the time evolution
\begin{equation}
\begin{aligned}
    &C: &&\begin{multlined}
        \frac{\diff }{\diff t} \theta(t)  = F(\theta(t))^{-1} \\
        \cdot \E_{\qrob} \bigg[ \mathcal L_{u(t)}^\dagger \big\{\1_{y(t)} \cdot 
       \nabla_{\theta(t)} \log \qrob_{\theta(t)}\big\}(X(t))\bigg].
    \end{multlined}
\end{aligned}
\label{eq:em_pred_c}
\end{equation}
For the reset condition, we have the same objective as in \cref{eq:em_update}, however, the reset is computed by minimizing the reverse \ac{kl} divergence \wrt 
\begin{equation*}
\begin{aligned}
    &C: &&\hat\pi_{t_i}(x) \coloneqq Z_i^{-1}  \1_{y_i}(x) [\mathcal L_{u(t)} \qrob_{\theta(t_i^-)}](x),
\end{aligned}
\label{eq:em_update_c}
\end{equation*}
with $Z_i =\sum_x  \1_{y_i}(x) [\mathcal L_{u(t)} \qrob_{\theta(t_i^-)}](x)$, for more see \citet{bronstein2018marginal}. For completeness we give the derivation of  \cref{eq:em_pred_d,eq:em_pred_c} in \cref{app:entropic}. We want to emphasize that these equations can also be derived from a geometrical approach, which is then known as projection filtering \citep{brigo1999approximate}.

The resulting equations for the time evolution in \cref{eq:em_pred_d,eq:em_pred_c} require only the solution of an \ac{ode} in the parameter space $\Theta$. Therefore, the computational complexity is only $d_{\theta}^2$ instead of $\vert \mathcal X \vert^2$.
Further, when assuming an exponential family parameterization, the optimal variational distribution at the discrete observation time points in \cref{eq:em_update} reduces to the problem of moment matching \citep{bishop2006pattern}. 

\section{\uppercase{Continuous-Time Control for POMDPs}}
Next, we describe how an optimal control trajectory $u_{[0,\infty)}^\ast$ can be found. 
First, we consider the problem of exact decision-making, which is however, intractable. 
Therefore, we present an approximate control method.
\subsection{Exact Control}\label{sec:opt_control}
We can describe the \ac{pomdp} by a belief \ac{mdp} using the filter distribution as the continuous state.
For the sake of presentation we assume for now, that the state space $\mathcal X$ is finite, therefore, we can represent the filtering distribution at time point $t$ as a vector $\pi_t \in \Delta^{\vert \mathcal X\vert}$, with components $\{\pi_t(x) \mid x \in \mathcal X\}$, where $\Delta^{\vert \mathcal X\vert}$ is the $\vert \mathcal X\vert$ dimensional probability simplex.
This allows us to use stochastic optimal control theory for continuous-valued states.
We define the value function for a belief $\pi \in \Delta^{\vert \mathcal X\vert}$ as the expected cumulative reward under the optimal control, \ie,
\begin{equation*}
\begin{aligned}
    &V(\pi) \\
    &\coloneqq \max_{u_{[t,\infty)}} \Eof*{\int_t^\infty \frac{1}{\tau} e^{-\frac{s-t}{\tau}} R(X(s),u(s))\diff s | \pi_t = \pi},   
\end{aligned}
\end{equation*}
where we use a normalization by $\frac{1}{\tau}$. 
Exploiting the principle of optimality, a Bellman equation can then be found in the form of \iac{hjb} equation \citep{alt2020pomdps} as
\begin{equation}
\begin{aligned}
    V(\pi)=\max_{u \in \mathcal U} &\E \bigg[R(X,u)+\tau \frac{\partial V(\pi)}{\partial \pi} f(\pi, u) \\
    & + \tau \lambda(X,u) (V(\pi+h(\pi,u))-V(\pi))\biggm| \pi\bigg],
\end{aligned}
\label{eq:belief_hjb}
\end{equation}
where the functions $f$, $h$ and $\lambda$ are defined by the filter dynamics, for more see the work of \citet{alt2020pomdps}.
An optimal policy can be retrieved by finding the maximizer of the \rhs of \cref{eq:belief_hjb}.
 
Hence, finding the optimal policy requires solving for the value function, which is generally hard, as it requires solving a $\vert \mathcal X\vert$-dimensional \ac{pde}. 
Therefore, even methods based on learning with function approximation, such as the one used in the work of \citet{alt2020pomdps} scale very poorly with the state-space size. 
Additionally, even when using approximate filter dynamics, as the ones previously discussed in \cref{sec:approx_inf}, the \rhs of the \ac{hjb} equation in \cref{eq:belief_hjb} requires solving expectations over the state, as well as the observation space.
This makes numerical solution methods, including learning-based \ac{pde} solution methods, intractable for all but very small state  and observation space problems.
For this reason, we present next an approximate control method, which does not require the solution of a high-dimensional \ac{pde}.

\subsection{Approximate Control} \label{sec:app_control}
Instead of computing the optimal control based on the dynamics of the filtering distribution we approximate it by separating the filtering and the control problem.
First, we compute the value function for the underlying \ac{mdp}.
We then combine it with the filtering distribution to approximate the value function of the \ac{pomdp} to retrieve a policy.
In the discrete-time literature, this is known as the Q\acs{mdp} method \citep{Littman95}, which has also recently found success using function approximation~\citep{karkus2017qmdp}.
Although the method assumes full observability for the planning and therefore does not consider the information gathering effect of actions, it leads to good results in many examples \citep{Littman95}.

Similar to the belief \ac{mdp} we define the value function of the underlying \ac{mdp} for a state $x \in \mathcal X$ as the expected cumulative reward under the optimal control, \ie,\looseness -1
\begin{equation*}
\begin{aligned}
    &V(x) \\
    &\coloneqq \max_{u_{[t,\infty)}} \Eof*{\int_t^\infty \frac{1}{\tau} e^{-\frac{s-t}{\tau}} R(X(s),u(s))\diff s | X(t) = x}, 
    \end{aligned}
\end{equation*}
where for the underlying \ac{mdp} we assume that the admissible control $u(t)$ can depend on the state $X(t)$.
By the principle of optimality the Bellman equation for continuous time and discrete state space  \citep{bertsekas2012adynamic,bertsekas2012bdynamic} is
\begin{equation}
    V(x)=\max_{u \in \mathcal U} R(x,u) + \tau \sum_{x' \neq x} \Lambda(x, x', u) (V(x') -  V(x)).
    \label{eq:hjb_state}
\end{equation}
The optimal policy for the \ac{mdp} maximizes the \rhs of \cref{eq:hjb_state}, which we define as state-action value function
\begin{equation*}
Q(x,u) \coloneqq R(x,u) + \tau \sum_{x' \neq x} \Lambda(x, x', u) (V(x') -  V(x)).
\end{equation*}
The definition of the state-action value function above can be reformulated as a contraction mapping, as
\begin{equation*}
\begin{multlined}
     Q(x,u) = \frac{R(x,u)}{1+\tau \sum_{x'\neq x} \Lambda(x,x',u)} \\
     +\tau \sum_{x' \neq x}
     \frac{ \Lambda(x,x',u)}{1+\tau \sum_{x'\neq x} \Lambda(x,x',u)} \max_{u'\in \mathcal U} Q(x',u').
\end{multlined}
\end{equation*}
The derivation is provided in \cref{app:mdp_control}.
When the state space is finite, above equation can be solved using fixed point iteration, similar to tabular dynamic programming methods like the value iteration algorithm in the discrete-time setting \citep{sutton2018reinforcement}.
For large or even infinite state spaces the fixed point iteration becomes intractable and we use value function approximation methods \citep{bradtke1994reinforcement}. For implementation details see \cref{app:Q_Learn}.

An approximate optimal policy for the \ac{pomdp} is then found as the maximizer of the expected state-action value function \ie,
\begin{equation}
    u(t)=\argmax_{u' \in \mathcal U} \E_{\qrob}[Q(X(t),u')],
    \label{eq:Q_control}
\end{equation}
where the expectation is \wrt the approximate filter distribution. 
For large or even infinite state spaces, where the expectation is intractable, we approximate it by Monte Carlo sampling to obtain a scalable policy.
\section{\uppercase{Experiments}} \label{sec:experiments}
To evaluate the efficacy of our method for partially observed systems we test it on three continuous-time discrete-state space control problems. 
We evaluate 
\begin{inlineitemize}
\item a controlled queueing network,
\item a controllable predator-prey system in form of a \ac{LV} model, and
\item a controlled closed-loop four species \ac{crn}.
\end{inlineitemize}\looseness-1
\subsection{Queueing Network}
\begin{figure}[ht]
    \centering
\includegraphics{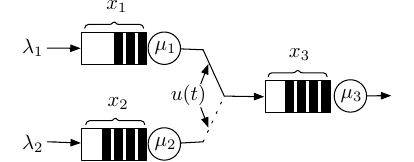}
    \caption{A schematic description of the considered queueing problem. The decision-maker decides which queue outputs its packets to the third queue.\looseness-1}
    \label{fig:Queueing_fig}
\end{figure}
\vfill
\begin{figure}
    \centering
    \includegraphics{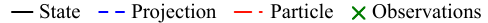}
    \includegraphics{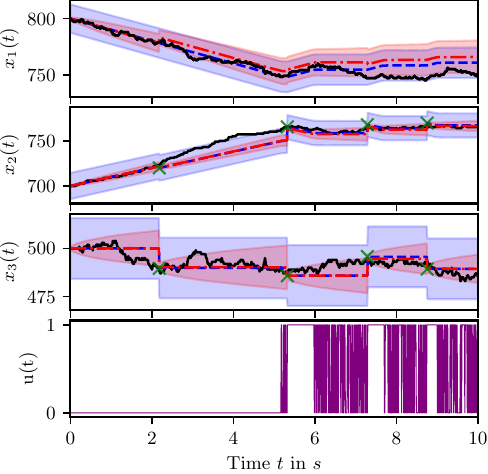}
    \caption{A sample trajectory of the queueing problem using a policy computed by the QMDP method. The upper plots compare the projection filter to a particle filter by indicating their mean and variance. The lower plot describes the actions over time.}
    \label{fig:Queueing_traj}
\end{figure}
First, we consider a queueing problem consisting of $n=3$ queues with fixed buffer size $N=1000$, \ie, $\mathcal{X}=\{0,1,\dots,N\}^n$.
The queues are connected as displayed in \cref{fig:Queueing_fig}.
Packets arrive with constant rates $\lambda_1$ and $\lambda_2$ in the first and the second queue, respectively. 
The action $u(t)$ decides which of the two queues (with service rates $\mu_1$ or $\mu_2$) preprocesses packets before sending them to the final queue. If $u(t) = 0$, queue $1$ is servicing queue $3$; if $u(t) = 1$, queue $2$ is servicing queue $3$.
The preprocessed packets are then being handled with service rate $\mu_3$.
As an observation model, we use Gaussian discrete-time measurements of queue 2 and 3.
The reward model is designed to favor empty queues. The parameters, reward function, observation model and additional information to all experiments can be found in  \cref{app:experiments}.

We use entropic matching to approximate the $(N+1)^n\approx 10^9$-dimensional exact filtering distribution by a binomial distribution for each queue
$
\qrob_{\theta}(x)=\prod_{i=1}^n \BinDis(x_i \mid N, \theta_i)
$,
with success probabilities $\{\theta_i\}_{i=1}^n$ and the total number of trials $N$ is fixed to the maximal buffer size.
Closed-form solutions for the drift of the binomial parameters for general queuing systems with finite buffer sizes are derived in \cref{app:buffer}.
The impact on the parameters by the measurements can be computed by \cref{eq:em_update}, which we approximate by moment matching right before and after the measurement to get closed-form updates, see \cref{app:buffer}.

The quality of the approximate filtering distribution plays a crucial role in the performance of our control method. To validate the effectiveness of our approach, we compare our projection filter against a bootstrap particle filter. Notably, the particle filter, configured with a sufficiently large number of samples $N_s = 10000$, serves as a robust benchmark or "ground truth" for our comparison.
\Cref{fig:Queueing_traj} shows the mean and variance of both filters for a sample trajectory.
The comparison highlights the strengths and the limitations of the projection filter. While it successfully captures the evolution of the mean induced by the prior dynamic, it struggles to capture the dynamics of the variance, due to the limited expressiveness of the parametric distribution. A second approximation error is noticeable at the observation times. As the distributions for each queue are modeled to be independent, there is no update in the first queue, when the other queues are being observed.
However, despite these limitations, the projection filter leads to an adequate approximation.

For all experiments, the policy is chosen according to our QMDP method. Therefore we learned the state action value function of the underlying MDP and approximate the expectation in \cref{eq:Q_control} using $k=20$ Monte Carlo samples. The resulting policy effectively maximizes the cumulative reward by activating the service rates of the queue where the filter expects more packets. When the beliefs of the first and second queue are equal, the policy frequently jumps between both actions, attempting to maintain the beliefs at the same level until a new observation updates them.\looseness -1

\subsection{Predator-Prey System}
Next, we consider the continuous-time discrete-state \ac{LV} model of \citet{wilkinson2018stochastic}.
The \ac{LV} problem consists of the following reactions, which are represented using the notation of \acp{crn} \citep{wilkinson2018stochastic} as
\begin{equation}
\begin{aligned}
&\mathsf X_1  \xrightarrow{c_1} 2 \mathsf X_1, &&\mathsf X_1 + \mathsf X_2  \xrightarrow{c_2} 2 \mathsf X_2, &&& \mathsf X_2  \xrightarrow{c_3(u(t))} \emptyset,
\end{aligned}
\label{eq:LV_model}
\end{equation}
where $\mathsf X_1$ is the prey species and $\mathsf X_2$ is the predator species. Hence, the unbounded state space of the system is $\mathcal X=\mathbb N_0^n$, with $n=2$. As an option to control the system, the decision maker can influence rate $c_3$ with $u(t) \in \mathcal{U}=\{0,1\}$, where $c_3(u(t)=1)=2 c_3(u(t)=0)$.
For the observation model, we use noisy discrete-time measurements of both states.
The reward model returns the negative Euclidean distance to a fixed target state $x^{\ast}=[100, 100]^\top$.
As an approximation to the infinite-dimensional filtering distribution we use a product Poisson distribution as
$
        \qrob_{\theta}(x)=\prod_{i=1}^n \PoisDis(x_i \mid \theta_i)
$.
\citet{bronstein2018variational} showed that using the entropic matching method with a product Poisson approximation leads to closed-form solutions for the drift of the parameters for general \acp{crn}. In \cref{app:poisson} we show generalize the derivation to include actions and provide details on \acp{crn}.\looseness-1
\begin{figure}[h]
    \centering
    \includegraphics{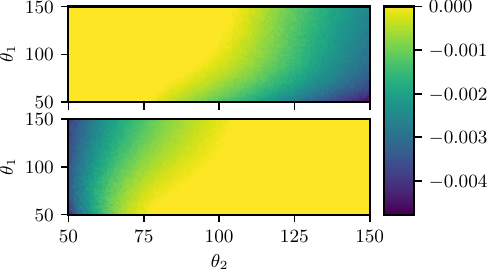}
    \caption{Advantage function for the \ac{LV} problem. The upper and the lower plot show the advantage function over a section of the
belief space for the first and the second action, respectively.}
    \label{fig:advantage}
\end{figure}

Using the Q\acs{mdp} method with the approximate filtering distribution we can compute the advantage function $A(\theta,u) \coloneqq\E_q[Q(X,u)-V(X)] $, visualized in \cref{fig:advantage}.
From this we can retrieve the policy as the actions that maximize the advantage function.

\begin{figure}[t]
    \centering
    \includegraphics{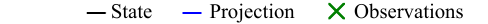}
    \includegraphics{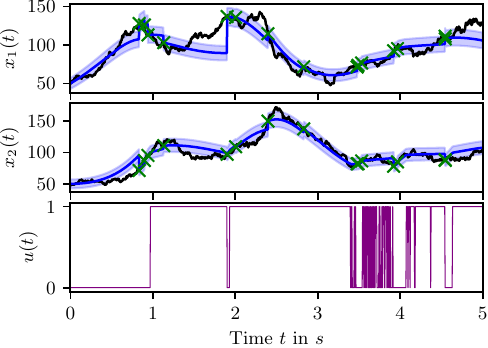}
    \caption{A sample trajectory of the LV problem using a policy computed by the QMPD method. The upper plots show the evolution of the exact states and the projection filter by indicating its mean and variance. The lower plot describes the actions over time.}
    \label{fig:LV_traj}
\end{figure}

\Cref{fig:LV_traj} depicts a sample trajectory of the controlled system. By comparison to samples with a constant control of either $u=0$ or $u=1$ (see \cref{app:experiments}), we see that the controlled trajectory is more stable in contrast to the oscillatory behaviour of the uncontrolled \ac{LV} problem. This demonstrates that the controller effectively combines the dynamics of both actions, resulting in trajectories that are closer to the goal state.

\subsection{Closed-Loop CRN}

 \begin{figure}[h]
 \centering
\includegraphics{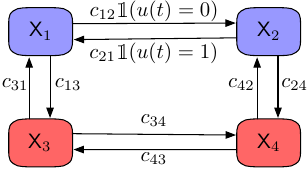}
    \caption{A schematic description of the considered \ac{crn}. Species $\mathsf{X_3}$ and $\mathsf{X_4}$ get observed exactly and are used to build an estimate of $\mathsf{X_1}$ and $\mathsf{X_2}$. The decision-maker can influence the flow between species $\mathsf{X_1}$ and $\mathsf{X_2}$.\looseness-1}
    \label{fig:mult_model}
\end{figure}
Finally, we use our method on a setup with sub-system continuous-time observations. 
We consider the $n=4$ species \ac{crn} described in \cref{fig:mult_model}.
The decision-maker controls the flow between species $\mathsf X_1$ and $\mathsf X_2$ with $u(t) = u$, where $u = 0$ corresponds to a flow from $\mathsf X_1$ to $\mathsf X_2$ and $u = 1$ corresponds to a flow from $\mathsf X_2$ to $\mathsf X_1$.
The state of the system is described by $X(t)=[\hat X^\top(t), \bar X^\top (t)]^\top \in \mathcal X$, where we consider that the first two species are unobserved, \ie, $\hat X(t)=[X_1(t),X_2(t)]^\top$ and the last two species are observed, \ie, $Y(t)=\bar X(t)=[X_3(t),X_4(t)]^\top$.
We design the reward function to favor a balanced system, where every species has the same number of molecules.
For closed-loop networks the total species number $N$ stays constant, therefore, we consider a filter approximation on $\mathcal X=\{0,1,\dots, N\}^n$ using the multinomial family over the states $\hat x$, with $x=[\bar x^\top, \hat x^\top]^\top$ as \looseness -1
\begin{equation*}
    \qrob_{\theta}(x) =\1_{y}(x) \MultDis(\hat x \mid N - \bar x_1 - \bar x_2, \theta).
\end{equation*}
\begin{figure}[t]
    \centering 
    \includegraphics{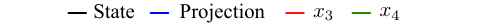}
    \includegraphics{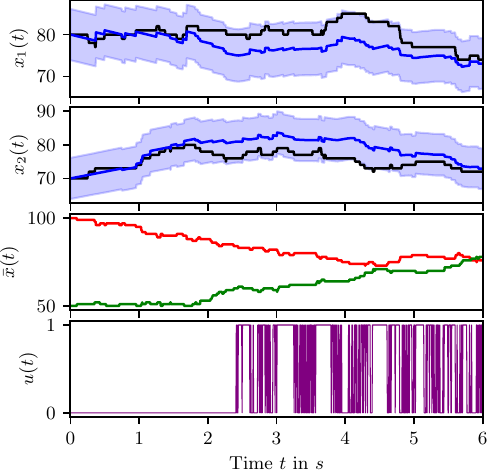}
    \caption{A sample trajectory of the \ac{crn} described in \cref{fig:mult_model} using a policy computed by the QMDP method. The upper plots show the evolution of the exact states and the filtering distribution by indicating its mean and variance. The lower plots describe the observed species and the actions over time. 
    }
    \label{fig:mult_traj}
\end{figure}In \cref{app:multinomial} we derive closed-form solutions for the drift and jump updates of the  multinomial parameters for general \acp{crn} with sub-system measurements using the method of entropic matching.

\Cref{fig:mult_traj} shows a sample trajectory of the \ac{crn} and the corresponding evolution of the filtering distribution. We can see that the filtering distribution captures the behaviour of the latent states successfully. Also, the resulting policy leads to reasonable decisions.
At the start of the trajectory, $x_3$ is the highest state while $x_4$ is the lowest. In response, the policy activates the flow from $x_1$ to $x_2$ to balance the state indirectly as a higher $x_2$ leads to a higher $x_4$ on average. In the second half of the trajectory the policy effectively switches between both actions to bring the system close to the goal state.

In \cref{app:experiments} we provide additional experiments for systems with smaller state spaces, where exact filtering is tractable, to evaluate the effect the filtering approximation has on the control performance.

\section{\uppercase{Discussion}}
To the best of our knowledge, there exists no other work addressing POMDPs in continuous time with large discrete state spaces. While this limits direct comparisons with existing methods, we view this as a distinctive strength of our work, providing a solid foundation that future research can build upon.

As an alternative to the entropic matching method, we considered the use of sequential Monte Carlo methods \citep{doucet2001sequential}, such as the particle filter, which we compared in the Queueing problem. However, while sequential Monte Carlo methods may be a viable option for approximate filtering, integrating them into a control method in the considered problem setting is far from trivial. 
The primary challenge lies in the control policy's reliance on the belief generated by all samples, which necessitates frequent updates when any individual sample changes.  While this update process is not an issue in discrete-time settings where all samples change simultaneously, it becomes impractically slow in continuous-time scenarios.
In contrast, the control method based on the entropic matching approach is straightforward and circumvents this issue.

\section{\uppercase{Conclusion}}\label{sec: conclusion}
In this work, we presented the optimal control theory for continuous-time \acp{pomdp} with discrete state and action spaces. We discussed the arising difficulties for the filtering and the control problem when dealing with large state spaces. For both problems, we described approximations that scale well with an increasing state space.
We then evaluated these methods on several partially observed systems.

In our experiments we approximated the filtering distributions with parametric distributions from the exponential family. 
While this choice allowed us to derive scalable closed-form solutions for the drift of the parameters, it is important to acknowledge that the expressiveness of these distributions has its limitations.
Looking ahead, future research can explore more advanced parametric families to further enhance the capabilities of our approach.
When it is not possible to obtain closed-form solutions, Monte Carlo methods offer an alternative approach to approximate the entropic matching equations.
Moreover, we are interested in applying the presented methods on \acp{pomdp}, where the latent dynamics are described by stochastic differential equations. In that scenario, the belief is in general infinite dimensional and could be approximated by continuous parametric families of distributions.
Also, this would enable the use for a broader range of applications including stochastic hybrid systems.

\subsubsection*{Acknowledgements}

This work has been co-funded by the Distr@l-project BlueSwarm (Project 71585164) of the Hessian Ministry of Digital Strategy and Development and the German Research Foundation (DFG) within the Collaborative Research Center (CRC) 1053 MAKI.

\bibliographystyle{abbrvnat}
\bibliography{bibliography}

\section*{Checklist}

 \begin{enumerate}

 \item For all models and algorithms presented, check if you include:
 \begin{enumerate}
   \item A clear description of the mathematical setting, assumptions, algorithm, and/or model. [Yes]
   \item An analysis of the properties and complexity (time, space, sample size) of any algorithm. [Yes, see the end of \cref{sec:approx_inf}]
   \item (Optional) Anonymized source code, with specification of all dependencies, including external libraries. [Yes]
 \end{enumerate}

 \item For any theoretical claim, check if you include:
 \begin{enumerate}
   \item Statements of the full set of assumptions of all theoretical results. [Yes]
   \item Complete proofs of all theoretical results. [Yes]
   \item Clear explanations of any assumptions. [Yes]     
 \end{enumerate}

 \item For all figures and tables that present empirical results, check if you include:
 \begin{enumerate}
   \item The code, data, and instructions needed to reproduce the main experimental results (either in the supplemental material or as a URL). [Yes]
   \item All the training details (e.g., data splits, hyperparameters, how they were chosen). [Yes]
         \item A clear definition of the specific measure or statistics and error bars (e.g., with respect to the random seed after running experiments multiple times). [Yes]
         \item A description of the computing infrastructure used. (e.g., type of GPUs, internal cluster, or cloud provider). [Not Applicable]
 \end{enumerate}

 \item If you are using existing assets (e.g., code, data, models) or curating/releasing new assets, check if you include:
 \begin{enumerate}
   \item Citations of the creator If your work uses existing assets. [Not Applicable]
   \item The license information of the assets, if applicable. [Not Applicable]
   \item New assets either in the supplemental material or as a URL, if applicable. [Not Applicable]
   \item Information about consent from data providers/curators. [Not Applicable]
   \item Discussion of sensible content if applicable, e.g., personally identifiable information or offensive content. [Not Applicable]
 \end{enumerate}

 \item If you used crowdsourcing or conducted research with human subjects, check if you include:
 \begin{enumerate}
   \item The full text of instructions given to participants and screenshots. [Not Applicable]
   \item Descriptions of potential participant risks, with links to Institutional Review Board (IRB) approvals if applicable. [Not Applicable]
   \item The estimated hourly wage paid to participants and the total amount spent on participant compensation. [Not Applicable]
 \end{enumerate}

 \end{enumerate}

\newpage
\appendix
\makeatletter
\renewcommand{\thesection}{\arabic{section}}
\makeatother

\onecolumn
\aistatstitle{Approximate Control for Continuous-Time POMDPs: \\
Supplementary Materials}

\section{\uppercase{Entropic Matching}}
\subsection{General}
\label{app:entropic}
In this subsection we give the derivation for the entropic matching equations \cref{eq:em_pred_d,eq:em_pred_c}, following the steps in \citet{bronstein2018marginal,bronstein2018variational}. 

As stated in \cref{sec:approx_inf}, we start by assuming a parametric form for the filtering distribution at time point $t$ as
\begin{equation*}
    \pi_t(x) \approx \qrob_{\theta(t)}(x),
\end{equation*}
with parameters $\theta(t) \in \Theta \subseteq \mathbb R^p$.
We then consider how the distribution $\qrob_{\theta(t)}(x)$ evolves in a small time step $h$ without new measurements.
For the observation model $D$, the evolution between observations is described in \cref{eq:pred_d}, this yields
\begin{equation*}\label{eq:pred_hstep}
    \tilde\pi_{t+h}(x) = \qrob_{\theta(t)}(x) + h [\mathcal L_{u(t)} \qrob_{\theta(t)}](x) + o(h).
\end{equation*}
The  parameters of the distribution are computed for a small time step $h$ by minimizing the reverse \ac{kl} divergence 
\begin{equation}\label{eq:KL_hstep}
    \theta(t+h)=\argmin_{\theta'} \KLof*{\tilde\pi_{t+h}| \qrob_{\theta'}}.
\end{equation}
The idea of the entropic matching method is to approximate the change of the filtering distribution in an infinitesimal time step such that the approximated distribution stays in the space of the parametric distributions. Therefore we first set \cref{eq:pred_hstep} in \cref{eq:KL_hstep}:

\begin{equation*}
\begin{multlined}
     \KLof*{\tilde\pi_{t+h}|\qrob_{\theta'}}\\
     \begin{aligned}
 &=\KLof*{\qrob_{\theta(t)} + h [\mathcal L_{u(t)} \qrob_{\theta(t)}] + o(h)|\qrob_{\theta'}}\\
     &
     \begin{multlined}
         =\sum_{x \in \mathcal X} \left(\qrob_{\theta(t)}(x) + h [\mathcal L_{u(t)} \qrob_{\theta(t)}](x) + o(h) \right)
     \cdot \log \frac{\qrob_{\theta(t)}(x) + h [\mathcal L_{u(t)} \qrob_{\theta(t)}](x) + o(h)}{\qrob_{\theta'}(x)}
     \end{multlined}\\
     &
     \begin{multlined}
          =\sum_{x \in \mathcal X} \left\{ \qrob_{\theta(t)}(x) \log \frac{\qrob_{\theta(t)}(x)}{\qrob_{\theta'}(x)} \right.
      \left.+ h  \left[ \qrob_{\theta(t)}(x) \frac{[\mathcal L_{u(t)} \qrob_{\theta(t)}](x)}{\qrob_{\theta(t)}(x)} +[\mathcal L_{u(t)} \qrob_{\theta(t)}](x) \log \frac{\qrob_{\theta(t)}(x)}{\qrob_{\theta'}(x)}\right] + o(h) \right\}
     \end{multlined}\\
     &
     \begin{multlined}
         =\KLof*{\qrob_{\theta(t)} | \qrob_{\theta'}}
     +h \E_{\qrob}\left[\frac{[\mathcal L_{u(t)} \qrob_{\theta(t)}](X(t))}{\qrob_{\theta(t)}(X(t))} + \frac{[\mathcal L_{u(t)} \qrob_{\theta(t)}](X(t))}{\qrob_{\theta(t)}(X(t))}\log \frac{\qrob_{\theta(t)}(X(t))}{\qrob_{\theta'}(X(t))}\right] +o(h),
     \end{multlined}\\
     \end{aligned}
     \end{multlined}
\end{equation*}
where we used a first order Taylor series around $h=0$ in the fourth line.
The \ac{kl} divergence between two members of a parametric distribution can be given by series expansion in $\theta -\theta'$ up to second order as
\begin{equation*}
    \KLof*{\qrob_{\theta} | \qrob_{\theta'}}=\frac{1}{2} (\theta'-\theta)^\top F(\theta) (\theta'-\theta),
\end{equation*}
where $F(\theta)\coloneqq - \E_{\qrob_{\theta}(x)}\left[\nabla_\theta \nabla_\theta^\top \log \qrob_{\theta}(X)\right]$ is the Fisher information matrix.
Hence, we can compute the minimum of \cref{eq:KL_hstep} as
\begin{equation*}
\begin{aligned}
        0&=\nabla_{\theta'} \KLof*{\tilde\pi_{t+h}(x)|\qrob_{\theta'}(x)}\vert_{\theta'=\theta(t+h)}\\
        &=F(\theta(t)) (\theta(t+h)-\theta(t))- h\E_{\qrob}\left[\nabla_{\theta(t+h)} \log \qrob_{\theta(t+h)}(X(t))\frac{[\mathcal L_{u(t)} \qrob_{\theta(t)}](X(t))}{\qrob_{\theta(t)}(X(t))}\right] +o(h).
\end{aligned}
\end{equation*}
Dividing both sides by $h$ and taking the limit $h \to 0$ we obtain
\begin{equation*}
    \begin{aligned}
        0&= F(\theta(t)) \frac{\diff}{\diff t} \theta(t) -\E_{\qrob}\left[ \nabla_{\theta(t)} \log \qrob_{\theta(t)}(X(t)) \frac{[\mathcal L_{u(t)} \qrob_{\theta(t)}](X(t))}{\qrob_{\theta(t)}(X(t))} \right]\\
        &=F(\theta(t)) \frac{\diff}{\diff t} \theta(t) - \sum_x \qrob_{\theta(t)}(x) \nabla_{\theta(t)} \log \qrob_{\theta(t)}(x) \frac{[\mathcal L_{u(t)} \qrob_{\theta(t)}](x)} {\qrob_{\theta(t)}(x)}\\
        &=F(\theta(t)) \frac{\diff}{\diff t} \theta(t) - \sum_x \nabla_{\theta(t)} \log \qrob_{\theta(t)}(x) [\mathcal L_{u(t)} \qrob_{\theta(t)}](x)\\
        &=F(\theta(t)) \frac{\diff}{\diff t} \theta(t) - \sum_x \qrob_{\theta(t)}(x) [\mathcal L_{u(t)}^\dagger   \nabla_{\theta(t)} \log \qrob_{\theta(t)}](x) \\
        &=F(\theta(t)) \frac{\diff}{\diff t} \theta(t) - \E_{\qrob}\left[\mathcal L_{u(t)}^\dagger   \nabla_{\theta(t)} \log \qrob_{\theta(t)}(X(t))\right].
        \end{aligned}
\end{equation*}

This leads to the entropic matching equation
 \begin{equation*}
     \frac{\diff}{\diff t} \theta(t)= F(\theta(t))^{-1} \E_{\qrob}\left[\mathcal L_{u(t)}^\dagger \nabla_{\theta(t)} \log \qrob_{\theta(t)}(X(t)) \right].
 \end{equation*}

For the sub-system measurement model $C$ in \cref{eq:model_c}, the evolution between between observation jumps is described in \cref{eq:em_pred_c}, this yields for the target filtering distribution
\begin{equation*}
          \tilde\pi_{t+h}(x) =\qrob_{\theta(t)}(x) + h\1_{y(t)}(x) [\mathcal L_{u(t)} \qrob_{\theta(t)}](x) 
    -h\qrob_{\theta(t)}(x)\sum_{x'} \1_{y(t)}(x') [\mathcal L_{u(t)} \qrob_{\theta(t)}](x') + o(h).
\end{equation*}

Setting this in \cref{eq:KL_hstep} leads to:
\begin{equation*}
\begin{multlined}
     \KLof*{\tilde\pi_{t+h}|\qrob_{\theta'}}\\
     \begin{aligned}
     &
     \begin{multlined}
         =\sum_{x \in \mathcal X} \left(\qrob_{\theta(t)}(x) + h\1_{y(t)}(x) [\mathcal L_{u(t)} \qrob_{\theta(t)}](x) 
    -h\qrob_{\theta(t)}(x)\sum_{x'} \1_{y(t)}(x') [\mathcal L_{u(t)} \qrob_{\theta(t)}](x') + o(h) \right)\\
     \cdot \left[\log \frac{\qrob_{\theta(t)}(x) + h\1_{y(t)}(x) [\mathcal L_{u(t)} \qrob_{\theta(t)}](x) 
    -h\qrob_{\theta(t)}(x)\sum_{x'} \1_{y(t)}(x') [\mathcal L_{u(t)} \qrob_{\theta(t)}](x') + o(h)}{\qrob_{\theta'}(x)}\right]
     \end{multlined}\\
     &
     \begin{multlined}
          =\sum_{x \in \mathcal X} \left\{ \qrob_{\theta(t)}(x) \log \frac{\qrob_{\theta(t)}(x)}{\qrob_{\theta'}(x)} \right.\\
      \left.+ h  \left[ \qrob_{\theta(t)}(x) \frac{\1_{y(t)}(x) [\mathcal L_{u(t)} \qrob_{\theta(t)}](x) -\qrob_{\theta(t)}(x)\sum_{x'} \1_{y(t)}(x') [\mathcal L_{u(t)} \qrob_{\theta(t)}](x')}{\qrob_{\theta(t)}(x)}\right.\right. \\
      +\left.\left.\left(\1_{y(t)}(x) [\mathcal L_{u(t)} \qrob_{\theta(t)}](x) 
    -\qrob_{\theta(t)}(x)\sum_{x'} \1_{y(t)}(x')[\mathcal L_{u(t)} \qrob_{\theta(t)}](x') \right) \log \frac{\qrob_{\theta(t)}(x)}{\qrob_{\theta'}(x)}\right] + o(h) \right\}
     \end{multlined}\\
     &
     \begin{multlined}
         =\KLof*{\qrob_{\theta(t)} | \qrob_{\theta'}}\\
     +h \E_{\qrob}\left[\frac{\1_{y(t)}(X(t)) [\mathcal L_{u(t)} \qrob_{\theta(t)}](X(t)) -\qrob_{\theta(t)}(X(t))\sum_{x'} \1_{y(t)}(x') [\mathcal L_{u(t)} \qrob_{\theta(t)}](x')}{\qrob_{\theta(t)}(X(t))}\right.\\
     \left.+ \frac{\1_{y(t)}(X(t)) [\mathcal L_{u(t)} \qrob_{\theta(t)}](X(t)) -\qrob_{\theta(t)}(X(t))\sum_{x'} \1_{y(t)}(x') [\mathcal L_{u(t)} \qrob_{\theta(t)}](x')}{\qrob_{\theta(t)}(X)}\log \frac{\qrob_{\theta(t)}(X(t))}{\qrob_{\theta'}(X(t))}\right] +o(h),
     \end{multlined}\\
     \end{aligned}
     \end{multlined}
\end{equation*}
where we used a first order Taylor series around $h=0$ in the third line.

By the same argument as before, we can compute the minimum of \cref{eq:KL_hstep} as
\begin{equation*}
\begin{aligned}
        0&=\nabla_{\theta'} \KLof*{\tilde\pi_{t+h}(x)|\qrob_{\theta'}(x)}\vert_{\theta'=\theta(t+h)}\\
    &=F(\theta(t)) (\theta(t+h)-\theta(t))\\
     &- h\E_{\qrob}\left[\nabla_{\theta(t+h)} \log \qrob_{\theta(t+h)}(X(t)) \frac{\1_{y(t)}(X(t)) [\mathcal L_{u(t)} \qrob_{\theta(t)}](X(t)) -\qrob_{\theta(t)}(X(t))\sum_{x'} \1_{y(t)}(x') [\mathcal L_{u(t)} \qrob_{\theta(t)}](x')}{\qrob_{\theta(t)}(X(t))}\right] +o(h)
\end{aligned}
\end{equation*}

Dividing both sides by $h$ and taking the limit $h \to 0$ we obtain
\begin{equation*}
    \begin{aligned}
        0&= F(\theta(t)) \frac{\diff}{\diff t} \theta(t) -\E_{\qrob}\left[ \nabla_{\theta(t)} \log \qrob_{\theta(t)}(X(t)) \frac{\1_{y(t)}(X(t)) [\mathcal L_{u(t)} \qrob_{\theta(t)}](X(t)) -\qrob_{\theta(t)}(X(t))\sum_{x'} \1_{y(t)}(x') [\mathcal L_{u(t)} \qrob_{\theta(t)}](x')}{\qrob_{\theta(t)}(X(t))} \right]\\
        &=F(\theta(t)) \frac{\diff}{\diff t} \theta(t) - \sum_x \qrob_{\theta(t)}(x) \nabla_{\theta(t)} \log \qrob_{\theta(t)}(x) \frac{\1_{y(t)}(x) [\mathcal L_{u(t)} \qrob_{\theta(t)}](x) -\qrob_{\theta(t)}(x)\sum_{x'} \1_{y(t)}(x') [\mathcal L_{u(t)} \qrob_{\theta(t)}](x')} {\qrob_{\theta(t)}(x)}\\
        &=F(\theta(t)) \frac{\diff}{\diff t} \theta(t) - \sum_x \nabla_{\theta(t)} \log \qrob_{\theta(t)}(x) \left(\1_{y(t)}(x) [\mathcal L_{u(t)} \qrob_{\theta(t)}](x) -\qrob_{\theta(t)}(x)\sum_{x'} \1_{y(t)}(x') [\mathcal L_{u(t)} \qrob_{\theta(t)}](x')\right)\\
                &=F(\theta(t)) \frac{\diff}{\diff t} \theta(t) - \sum_x \nabla_{\theta(t)} \log \qrob_{\theta(t)}(x) \left(\1_{y(t)}(x) [\mathcal L_{u(t)} \qrob_{\theta(t)}](x) \right)\\
        &=F(\theta(t)) \frac{\diff}{\diff t} \theta(t) - \sum_x \qrob_{\theta(t)}(x) \mathcal L_{u(t)}^\dagger \left\{\1_{y(t)} \cdot  \nabla_{\theta(t)}  \log \qrob_{\theta(t)}\right\}(x) \\
        &=F(\theta(t)) \frac{\diff}{\diff t} \theta(t) - \E_{\qrob}\left[\mathcal L_{u(t)}^\dagger\left\{ \1_{y(t)}\cdot  \nabla_{\theta(t)} \log \qrob_{\theta(t)}\right\}(X(t))\right],
        \end{aligned}
\end{equation*}
where in the fourth line we used 
 $\E_{\qrob}\left[\nabla_{\theta(t)} \log \qrob_{\theta(t)}(X(t))\right]=0$.
 This leads to the entropic matching equation
 \begin{equation*}
     \frac{\diff}{\diff t} \theta(t)= F(\theta(t))^{-1} \E_{\qrob}\left[\mathcal L_{u(t)}^\dagger\left\{ \1_{y(t)}\cdot  \nabla_{\theta(t)} \log \qrob_{\theta(t)}\right\}(X(t))\right].
 \end{equation*}

\subsection{Entropic Matching for Finite Buffer Queues}
\label{app:buffer}
We consider a  queueing network  with $n$ queues.
We assumed that the $i$the queue is an $M/M/c/n$ queue, with Markovian arrivals and services, a number of $c$ servers and a finite buffer of size $n$, for more, see \citep{bolch2006queueing}.
The state of the queueing network is described by the number of packets in the queues, i.e., $x=[x_1,\dots, x_n]^\top$, with $x_i \in \mathcal X_i \coloneqq \{0,1,\dots, N_i\}$ and $\mathcal X \coloneqq \bigtimes_{i=1}^n \mathcal X_i$. 
We denote the stochastic routing matrix by $\mathbf P \in \Delta^{n+1\times n+1}$. 
The entry $P_{ij}$ denotes the probability of routing a packet from queue $i$ to queue $j$, with $(i,j) \in \{1,\dots n\} \times \{1,\dots,n\}$.
Additionally, we denote by the entries $P_{i,n+1}$ and $P_{n+1, i}$ the probabilities that a packet leaves and enters the queueing network from and into the queue $i$, respectively.
The corresponding arrival, and respectively service, rates are denoted by $\{\tilde \lambda_{ij}\}$. 
The effective rate for a number of $c_i$ servers is then given by 
\begin{equation*}
    \lambda_{ij}(x_i)=\tilde \lambda_{ij} P_{ij} \min(x_i, c_i),
\end{equation*}
where we set $x_{n+1}=c_{n+1}=1$ and $\mathcal X_{n+1}=\mathbb N_0$ for convenience.
We define the corresponding jump vectors as
\begin{equation*}
    \nu_{ij} \coloneqq e_j-e_i,
\end{equation*}
where $e_i$ is the $i$th unit vector, and we set by definition $e_{n+1}$ to the $n$-dimensional zero vector.
The system is then described by a continuous-time Markov chain $\{X(t)\}_{t \in \mathbb R_{\geq 0}}$, with reaction rate function $\Lambda(x,x') \coloneqq \lim_{h \to 0} h^{-1} \Prob(X(t+h)=x'\mid X(t)=x)$ given as
\begin{equation*}
   \Lambda(x,x')= \1(x'\in \mathcal X) \1(x\in \mathcal X)\sum_{i=1}^{n+1} \sum_{j=1}^{n+1} \lambda_{ij}(x_i) \1(x'=x+ \nu_{ij}) ,
\end{equation*}
for all $(x,x') \in \mathcal X \times \mathcal X$.
The forward master equation is therefore given by 
\begin{equation*}
\begin{aligned}
    \frac{\mathrm d}{\mathrm d t} \prob_t(x)&= [\mathcal L \prob_t](x)\\
    &=\sum_{x'} \Lambda(x',x) \prob_t(x')\\
    &=\sum_{x' \neq x} \Lambda(x',x) \prob_t(x')-\sum_{x' \neq x} \Lambda(x,x') \prob_t(x)\\
    &=\sum_{x' \neq x} \sum_{i=1}^{n+1} \sum_{j=1}^{n+1} \lambda_{ij}(x_i')   \1(x=x'+ \nu_{ij})\prob_t(x') 
    -\sum_{x' \neq x} \sum_{i=1}^{n+1} \sum_{j=1}^{n+1} \lambda_{ij}(x_i)  \1(x'=x+ \nu_{ij})\prob_t(x)\\
    &=\sum_{i=1}^{n+1} \sum_{j=1}^{n+1} \lambda_{ij}(x_i + 1) \1(x- \nu_{ij} \in \mathcal X) \prob_t(x - \nu_{ij})
     - \sum_{i=1}^{n+1} \sum_{j=1}^{n+1} \lambda_{ij}(x_i) \1(x + \nu_{ij} \in \mathcal X) \prob_t(x)\\
    \frac{\mathrm d}{\mathrm d t} \prob_t(x)&= \sum_{i=1}^{n+1} \sum_{j=1}^{n+1} \lambda_{ij}(x_i+1) \1(x_i+1 \in \mathcal X_i)\1(x_j-1 \in \mathcal X_j)\prob_t(x - \nu_{ij})\\
    &\quad - \sum_{i=1}^{n+1} \sum_{j=1}^{n+1} \lambda_{ij}(x_i) \1(x_i - 1 \in \mathcal X_i)\1(x_j + 1 \in \mathcal X_j) \prob_t(x).
    \end{aligned}
\end{equation*}
Let, $\mathcal L^\dagger$ denote the adjoint operator of $\mathcal L$, w.r.t, the inner product 
$\langle \phi, \psi\rangle \coloneqq \sum_{x} \phi(x) \psi(x)$, i.e,
\begin{equation*}
\begin{aligned}
      &\langle \mathcal L \phi, \psi\rangle = \langle \mathcal \phi, [\mathcal L^\dagger \psi]\rangle\\
      \Longleftrightarrow &  \sum_{x} [\mathcal L \phi](x) \psi(x) = \langle \mathcal \phi, [\mathcal L^\dagger \psi]\rangle\\
       \Longleftrightarrow &  \sum_{x} \sum_{x'} \Lambda(x',x)\phi(x') \psi(x) =  \langle \mathcal \phi, [\mathcal L^\dagger \psi]\rangle\\
       \Longleftrightarrow &   \sum_{x'} \phi(x') \sum_{x} \Lambda(x',x) \psi(x) = \langle \mathcal \phi, [\mathcal L^\dagger \psi]\rangle.\\
\end{aligned}
\end{equation*}
Hence, we have
\begin{equation*}
    [\mathcal L^\dagger \phi](x) = \sum_{x'} \Lambda(x,x') \phi(x').
\end{equation*}

Therefore, we can find the adjoint operator $\mathcal L^\dagger$ working on a test function $\phi$ as
\begin{equation*}
    \begin{aligned}
        [\mathcal L^\dagger \phi] (x) &= \sum_{x'} \Lambda(x,x') \phi(x') \\
        &=\sum_{x'\neq x} \Lambda(x,x') \phi(x')  - \sum_{x'\neq x} \Lambda(x,x') \phi(x)\\
        &=\sum_{x'\neq x} \Lambda(x,x') (\phi(x') -\phi(x))\\
        &=\sum_{x'\neq x} \sum_{i=1}^{n+1} \sum_{j=1}^{n+1} \lambda_{ij}(x_{i}) \1(x'=x+ \nu_{ij})  (\phi(x') -\phi(x))\\
        [\mathcal L^\dagger \phi] (x) &=\sum_{i=1}^{n+1} \sum_{j=1}^{n+1} \lambda_{ij}(x_{i}) \1(x_i-1 \in \mathcal X_i) \1(x_j+1 \in \mathcal X_j)  (\phi(x+ \nu_{ij}) -\phi(x)).    
    \end{aligned}
\end{equation*}
Let, $\phi(x)=\nabla \log \qrob_{\theta}(x)$ and $\qrob_{\theta}(x)= \prod_{i=1}^n \BinDis(x_i \mid N_i, \theta_i)$, we compute
\begin{equation*}
    \partial_{\theta_i} \log \qrob_{\theta}(x)=\frac{x_i-N_i\theta_i}{\theta_i (1-\theta_i)}.
\end{equation*}
The inverse Fisher information matrix is given by
\begin{equation*}
    F(\theta)^{-1}=\diag\left(\left[\frac{\theta_1(1-\theta_1)}{N_1},\dots,\frac{\theta_n(1-\theta_n)}{N_n}\right]^\top\right).
\end{equation*}
Note that,
\begin{equation*}
     \nabla_{\theta} \log \qrob_{\theta}(x + \nu_{ij}) -  \nabla_{\theta} \log \qrob_{\theta}(x)=\frac{e_j}{\theta_j (1-\theta_j)}-\frac{e_i}{\theta_i (1-\theta_i)}.
\end{equation*}
Therefore we have for the entropic matching equation
\begin{equation*}
\begin{aligned}
     \frac{\mathrm d }{\mathrm d t} \theta &= F(\theta)^{-1}\E_{\qrob}[\mathcal L^\dagger \nabla_{\theta} \log \qrob_{\theta}(X)],
\end{aligned}
\end{equation*}
the ODE
\begin{equation*}
\begin{aligned}
     \frac{\mathrm d }{\mathrm d t} \theta(t) &= \E_{\qrob}\left[\sum_{i=1}^{n+1}   \sum_{j=1}^{n+1} \lambda_{ij}(X_{i}) \1(X_i-1 \in \mathcal X_i) \1(X_j+1 \in \mathcal X_j)(\frac{e_j}{N_j}-\frac{e_i}{N_i}) \right]\\
     &=\sum_{i=1}^{n+1}  \sum_{j=1}^{n+1} (\frac{e_j}{N_j}-\frac{e_i}{N_i})  \E_{\qrob}\left[\lambda_{ij}(X_{i}) \1(X_i-1 \in \mathcal X_i)] \E[\1(X_j+1 \in \mathcal X_j) \right].
     \end{aligned}
\end{equation*}
This can be written component-wise as 
\begin{equation*}
\begin{aligned}
 \frac{\mathrm d }{\mathrm d t} \theta_i(t)=&\frac{1}{N_i} \sum_{j=1}^{n+1}\left( \E_{\qrob}[\lambda_{ji}(X_{j}) \1(X_j-1 \in \mathcal X_j)] \E_{\qrob}[\1(X_i+1 \in \mathcal X_i) ]\right.\\
 &\left.- \E_{\qrob}[\lambda_{ij}(X_{i}) \1(X_i-1 \in \mathcal X_i)] \E_{\qrob}[\1(X_j+1 \in \mathcal X_j) ]\right).
\end{aligned}
\end{equation*}
We compute for $j \in \{1,\dots, n\}$
\begin{equation*}
\begin{aligned}
 \E_{\qrob}[\1(X_j+1 \in \mathcal X_j) ]&=\sum_{x_j=0}^{N_j} \BinDis(x_j \mid N_j, \theta_j)  \1(x_j+1 \in \mathcal X_j)\\
 &=1-\sum_{x_j=0}^{N_j} \BinDis(x_j \mid N_j, \theta_j)  \1(x_j+1 \notin \mathcal X_j)\\
 &=1- \BinDis(N_j \mid N_j, \theta_j)
     \end{aligned}
\end{equation*}
and for $j=n+1$
\begin{equation*}
 \E_{\qrob}[\1(X_j+1 \in \mathcal X_j) ]=1.
 \end{equation*}
For $i\in \{1,\dots, n\}$
\begin{equation*}
    \begin{aligned}
         \E_{\qrob}[\lambda_{ij}(X_{i}) \1(X_i-1 \in \mathcal X_i)]&=\sum_{x_i=0}^{N_i} \BinDis(x_i \mid N_i, \theta_i) \tilde \lambda_{ij} P_{ij} \min(x_{i},c_i) \1(x_i-1 \in \mathcal X_i)\\
         &=\tilde \lambda_{ij} P_{ij} (\sum_{x_i=0}^{c_{i}-1} \BinDis(x_i \mid N_i, \theta_i) x_i + \sum_{x_i=c_i}^{N_i}  \BinDis(x_i \mid N_i , \theta_i) c_i)\\
         &= \tilde \lambda_{ij} P_{ij} (\sum_{x_i=0}^{c_{i}-1} \BinDis(x_i \mid N_i, \theta_i) x_i +c_i(1- \sum_{x_i=0}^{c_i-1}  \BinDis(x_i \mid N_i, \theta_i))\\
         &=\tilde \lambda_{ij} P_{ij} (c_i + \sum_{x_i=0}^{c_{i}-1} \BinDis(x_i \mid N_i, \theta_i) (x_i-c_i))
    \end{aligned}
\end{equation*}
and for $i=n+1$ we have
\begin{equation*}
    \E_{\qrob}[\lambda_{ij}(X_{i}) \1(X_i-1 \in \mathcal X_i)]=\tilde \lambda_{ij} P_{ij}.
\end{equation*}
Therefore, we have the component-wise ODE
\begin{equation*}
\begin{aligned}
    \frac{\mathrm d }{\mathrm d t} \theta_i(t)&=\frac{1}{N_i} \sum_{j=1}^{n+1} \E_{\qrob}[\lambda_{ji}(X_{j}) \1(X_j-1 \in \mathcal X_j)] \E_{\qrob}[\1(X_i+1 \in \mathcal X_i) ]\\
    &\qquad- \E_{\qrob}[\lambda_{ij}(X_{i}) \1(X_i-1 \in \mathcal X_i)] \E_{\qrob}[\1(X_j+1 \in \mathcal X_j) ]\\
    &=\frac{1}{N_i}\sum_{j=1}^{n} (\tilde \lambda_{ji} P_{ji} (c_j + \sum_{x_j=0}^{c_{j}-1} \BinDis(x_j \mid N_j, \theta_j) (x_j-c_j))(1- \BinDis(N_i \mid N_i, \theta_i))\\
    &\qquad-\tilde \lambda_{ij} P_{ij} (c_i + \sum_{x_i=0}^{c_{i}-1} \BinDis(x_i \mid N_i, \theta_i) (x_i-c_i))(1- \BinDis(N_j \mid N_j, \theta_j))\\
    &\qquad +\frac{1}{N_i} \tilde \lambda_{n+1,i} P_{n+1,i}(1- \BinDis(N_i \mid N_i, \theta_i))- \frac{1}{N_i}\tilde \lambda_{i,n+1} P_{i,n+1} (c_i + \sum_{x_i=0}^{c_{i}-1} \BinDis(x_i \mid N_i, \theta_i) (x_i-c_i))\\
    &=\frac{1- \BinDis(N_i \mid N_i, \theta_i)}{N_i}(\tilde \lambda_{n+1,i} P_{n+1,i}+\sum_{j=1}^{n} \tilde \lambda_{ji} P_{ji} (c_j + \sum_{x_j=0}^{c_{j}-1} \BinDis(x_j \mid N_j, \theta_j) (x_j-c_j)))\\
    &-\frac{c_i + \sum_{x_i=0}^{c_{i}-1} \BinDis(x_i \mid N_i, \theta_i) (x_i-c_i)}{N_i}(\tilde \lambda_{i,n+1} P_{i,n+1}+\sum_{j=1}^{n} \tilde \lambda_{ij} P_{ij}(1- \BinDis(N_j \mid N_j, \theta_j)))
\end{aligned}
\end{equation*}
In the example we assume that the number of servers is set to $c_1=c_2=c_3=1$.
Note that, we have $\BinDis(0\mid N_i, \theta_i)=(1-\theta_i)^{N_i}$ and  $\BinDis(N_i\mid N_i, \theta_i)=\theta_i^{N_i}$, hence,
\begin{equation*}
\begin{aligned}
    \frac{\mathrm d }{\mathrm d t} \theta_i&=\frac{1-\theta_i^{N_i}}{N_i}(\tilde \lambda_{n+1,i} P_{n+1,i}+\sum_{j=1}^{n} \tilde \lambda_{ji} P_{ji} (1- (1-\theta_j)^{N_j})-\frac{1- (1-\theta_i)^{N_i}}{N_i}(\tilde \lambda_{i,n+1} P_{i,n+1}+\sum_{j=1}^{n} \tilde \lambda_{ij} P_{ij}(1- \theta_j^{N_j})).
    \end{aligned}
\end{equation*}
In the experiments we use a queueing network with $n=3$, queues with equal buffer size $N=N_1=N_2=N_3$ and  We set the routing matrix to 
\begin{equation*}
 \mathbf P = \begin{bmatrix}
     & 0 & 0 & \1(u(t)=0) & 0\\
     & 0 & 0 & \1(u(t)=1) & 0\\
     & 0 & 0 & 0 & 1\\
     & 1 & 1 & 0 & 0
 \end{bmatrix}.
\end{equation*}

At observation time points we update the parameters accordingly as
\begin{equation*}
    \theta(t_i)= \argmin_{\theta'} \KLof*{\pi_{t_i\mid \theta(t_i^-)} | \qrob_{\theta'} },
\end{equation*}
where the \ac{kl} divergence is computed \wrt 
\begin{equation*}
\begin{aligned}
    \pi_{t_i\mid \theta(t_i^-)}(x) \propto  \prob(y_i \mid x , u(t_i^-)) \qrob_{\theta(t_i^-)}(x)).
\end{aligned}
\end{equation*}

For members of the exponential family parameterization as 
\begin{equation*}
    \qrob_{\theta}(x)=q_0(x)\exp(\theta^\top T(x)-A(\theta)),
 \end{equation*}
with base measure $q_0(x)$, natural parameters $\theta$, sufficient statistics $T(x)$, and log-normalizer $A(\theta)$, the optimization at the discrete observation time points, reduces to the problem of moment matching \citep{bishop2006pattern}, \ie,
\begin{equation*}
   \E_{\qrob}[T(X)]=\E_{\pi}[T(X)].
   \label{eq:em_moment_matching}
\end{equation*}
While this can be computed exact for examples with small state spaces, it can be approximated with Monte Carlo samples for large state spaces. In our experiments we use a different approximation for the update in order to obtain closed-form solutions. For queues with a large buffer size $N_i$, we approximate the binomial distribution at observation times by a Gaussian distribution via matching the moments, \ie, $\mathcal{N}(x_i\mid N_i\theta_i,N_i\theta_i(1-\theta_i))$. Since we consider Gaussian measurements, we can compute the posterior distribution and again match the moments to approximate the posterior by a binomial distribution.

\subsection{Entropic Matching for Chemical Reaction Networks using a Product Poisson Distribution} \label{app:poisson}

Chemical reaction networks \citep{wilkinson2018stochastic} are a subclass of \acp{ctmc} defined as a system of $k$ reactions involving $n$ species as
\begin{align*}
    S_{11}\mathsf{X}_1 + \dots + S_{n1} \mathsf{X}_n  &\xrightarrow{c_1} P_{11}\mathsf{X}_1 + \dots + P_{n1} \mathsf{X}_n \\
    &\vdots\\
        S_{1k}\mathsf{X}_1 + \dots +  S_{nk} \mathsf{X}_n  &\xrightarrow{c_k} P_{1k}\mathsf{X}_1 + \dots + P_{nk} \mathsf{X}_n,   
\end{align*}
where $c_j \in \mathbb R_{\geq 0 }$ is called the reaction rate of the $j$-th reaction and $S_{ij} \in \mathbb N_{0}$ and $P_{ij} \in \mathbb N_{0}$  are the stoichiometric substrate and product coefficients for species $\mathsf{X_i}$ in the $j$-th reaction, respectively. 
The state of the network is described by the size of each species, \ie, $x=[x_1,\dots, x_n]^\top$, with $x_i \in \mathcal X_i \coloneqq \mathbb N_{0}$ and $\mathcal X \coloneqq \bigtimes_{i=1}^n \mathcal X_i$. 
The change vector $v_j \in \mathbb Z^n$ corresponding to the $j$-th reaction is defined by
\begin{align*}
    v_j = \begin{pmatrix}
        P_{1j} - S_{1j}\\
         \vdots \\
         P_{nj} - S_{nj}
    \end{pmatrix},
\end{align*}
and the propensity corresponding to the $j$-th reaction is given by mass-action kinetics as
\begin{equation*}
    \lambda_j(x,u)= c_j(u) \prod_{i=1}^{n}  \binom{x_i}{S_{ij}}.
\end{equation*}
Note that we add the possibility to control the \ac{crn} by making the reaction coefficients action dependent.
From this we can define the rate function of the \ac{ctmc} as
\begin{equation*}
    \Lambda(x,x',u) = \sum_{j=1}^k \1(x'=x+v_j)\lambda_j(x,u).
\end{equation*}

We now want to approximate the filtering distribution of a \ac{crn} with discrete time observations by a product Poisson distribution as
\begin{equation*}
        \qrob_{\theta}(x)=\prod_{i=1}^n \PoisDis(x_i \mid \theta_i).
\end{equation*}
Using the entropic matching method the evolution of the parameters can be described by
\begin{equation*}
    \frac{\diff }{\diff t} \theta(t) = F(\theta(t))^{-1} \E_{\qrob} \left[\mathcal L_{u(t)}^\dagger \nabla_{\theta(t)} \log \qrob_{\theta(t)}(X(t))\right].
\end{equation*}

For a set of reactions the adjoint operator of the evolution operator which acts on functions $\psi$ is given by
\begin{equation*}
    [\mathcal L_{u}^{\dagger}\psi](x)=\sum_{j=1}^k\lambda_j(x,u)\{\psi(x+v_j)-\psi(x)\}.
\end{equation*}
As $\nabla_{\theta(t)} \log \qrob_{\theta(t)}(x)$ is linear in $x$, the term inside the expectation can be reduced to
\begin{align*}
    [\mathcal L_{u(t)}^{\dag} \nabla_{\theta(t)} \log \qrob_{\theta(t)}](x) = \sum_{j=1}^k \frac{v_j}{\theta} \lambda_j(x,u(t)).
\end{align*}

Combining this with the Fisher matrix of the product Poisson distribution the drift of the $l$th parameter simplifies to
\begin{align*}
    \frac{\diff}{\diff t}  \theta_l(t) &= \theta_l(t) \E_{\qrob} \left[\frac{1}{\theta_l(t)}\sum_{j=1}^k\lambda_j(X(t),u(t)) v_{lj}\right]\\
    & = \sum_{j=1}^k\E_{\qrob}\left[{\lambda_j(X(t),u(t))}\right]v_{lj}\\
    & = \sum_{j=1}^k\E_{\qrob}\left[c_j(u(t))\prod_{i=1}^n {\binom{X_i(t)}{S_{ij}}}\right]v_{lj}\\
        & = \sum_{j=1}^kc_j(u(t))\prod_{i=1}^n\E_{\qrob}\left[ \frac{X_i(t)!}{S_{ij}!(X_i(t)-S_{ij})!}\right]v_{lj}\\
                & = \sum_{j=1}^kc_j(u(t))\prod_{i=1}^n\frac{\theta_i^{S_{ij}}(t)}{S_{ij}!}v_{lj}.
\end{align*}
For the \ac{LV} problem considered in the experiments we get the following drift:
\begin{equation*}
    \begin{aligned}
            \frac{\diff}{\diff t} \theta_1(t)&=c_1\theta_1(t)-c_2\theta_1(t)\theta_2(t)+c_1\\
    \frac{\diff}{\diff t} \theta_2(t)&=c_{2}\theta_1(t)\theta_2(t)-c_{3}(u(t))\theta_2(t).
    \end{aligned}
\end{equation*} 
Combining the drift in between measurements with the moment matching method at the time points new measurements are given, we can describe the evolution of the parameters in time.

Similar to the entropic matching for finite buffer queues using a product binomial distribution, we can compute the update at observation time points using moment matching, which can be approximated with Monte Carlo samples. In our experiments we use a different approximation for the update in order to obtain closed-form solutions. We approximate the Poisson distribution at observation times by a Gaussian distribution via matching the moments, \ie, $\mathcal{N}(x_i\mid \theta_i,\theta_i)$. Since we consider Gaussian measurements, we can compute the posterior distribution and again match the moments to approximate the posterior by a Poisson distribution.

\subsection{Entropic Matching for Chemical Reaction Networks with Sub-System Measurements using a Multinomial Distribution}
\label{app:multinomial}
Consider \acp{crn} with $n=\hat n+ \bar n$ species.
We denote the state of all species in the system as $x=[\hat x^\top, \bar x^\top]^\top$, where the first $\hat n$ states, also denoted by $\hat{x}=[x_1,\dots, x_{\hat n}]^\top$, are being estimated using exact continuous observations of the latter $\bar n$ states $\bar x=[x_{\hat n+1},x_{\hat n+2},\dots,x_{\hat n + \bar n}]^\top$.
The exact filter for these systems is considered in \citep{bronstein2018marginal}.

We want to approximate the exact filter using a multinomial distribution as
\begin{equation*}
\qrob_{\theta}(x)=\1_{y(t)}(x) \qrob_{\theta}(\hat x) = \1_{y(t)}(x) \MultDis(\hat x \mid N - \sum_{i=1}^{\bar n} \bar x_i, \theta),
\end{equation*}
with event probabilities $\theta$ and number of trials $\hat{N}=N-\sum_{i=1}^{\bar n} \bar x_i$.

Using the method of entropic matching we compute the evolution of the parameters as
\begin{align*}
    \frac{\mathrm d }{\mathrm d t} \theta(t)
    &= F(\theta(t))^{-1} \E_{\qrob}\left[\mathcal L_{u(t)}^{\dag}\{\1_{y(t)}\cdot\nabla_{\theta(t)} \log \qrob_{\theta(t)}\}(X(t))\right] \\
    &= F(\theta(t))^{-1} \E_{\qrob}\left[\sum_{j=1}^Rh_j(X(t))\{\1_{y(t)}(X(t)+v_j)\nabla_{\theta(t)} \log \qrob_{\theta(t)}(X(t)+v_j)\right.\\
    &\left.\qquad- \1_{y(t)}(X(t))\nabla_{\theta(t)} \log \qrob_{\theta(t)}(X(t))\}\right]\\
    &= F(\theta(t))^{-1} \E_{\qrob}\left[\sum_{j=1}^Rh_j(X(t))\{\1(\Bar{v}_j=0)(\nabla_{\theta(t)} \log \qrob_{\theta(t)}(X(t)+v_j)-\nabla_{\theta(t)} \log \qrob_{\theta(t)}(X(t)))\right.\\
    &\left.\qquad-\1(\Bar{v}_j\neq 0)\nabla_{\theta(t)} \log \qrob_{\theta(t)}(X(t))\}\right]\\
      &= F(\theta(t))^{-1} \sum_{j=1}^Rc_j\prod_{i=1}^{\Bar{n}}\binom{y_i}{\Bar{S}_{ij}}\\
      &\qquad\left(\1(\Bar{v}_j=0) \E_{\qrob}\left[\prod_{i=1}^{\Hat{n}}\binom{\Hat{X}_i(t)}{\Hat{S}_{ij}}(\nabla_{\theta(t)} \log \qrob_{\theta(t)}(X(t)+v_j)-\nabla_{\theta(t)} \log \qrob_{\theta(t)}(X(t)))\right]\right.\\
    &\qquad\left.-\1(\Bar{v}_j\neq 0)\E_{\qrob}\left[\prod_{i=1}^{\Hat{n}}\binom{\Hat{X}_i(t)}{\Hat{S}_{ij}}\nabla_{\theta(t)} \log \qrob_{\theta(t)}(X(t))\right]\right),  
\end{align*}

where the gradient of the log probability is given by
\begin{align*}
  \nabla_{\theta(t)}  \log \qrob_{\theta(t)}(x) = \1_{y(t)}(x)\begin{pmatrix} \frac{\hat{x}_1}{\theta_1(t)} - \frac{\hat{x}_{\hat{n}}}{\theta_{\hat{n}}(t)}\\
   \vdots\\
   \frac{\hat{x}_{\hat{n}-1}}{\theta_{\hat{n}-1}(t)}-\frac{\hat{x}_{\hat{n}}}{\theta_{\hat{n}}(t)}
   \end{pmatrix}.
\end{align*}

This leads to 

\begin{align*}
   \frac{\mathrm d }{\mathrm d t}\theta(t) &= 
 F(\theta(t))^{-1} \sum_{j=1}^Rc_j\prod_{i=1}^{\Bar{n}}\binom{y_i}{\Bar{S}_{ij}}\\
&\qquad\left(\1(\Bar{v}_j=0) \E_{\qrob}\left[\prod_{i=1}^{\Hat{n}}\binom{\Hat{X}_i(t)}{\Hat{S}_{ij}}\right]\begin{pmatrix} \frac{\hat{v}_{1j}}{\theta_1(t)} - \frac{\hat{v}_{\hat{n}j}}{\theta_{\hat{n}}(t)}\\
   \vdots\\
   \frac{\hat{v}_{\hat{n}-1,j}}{\theta_{\hat{n}-1}(t)}- \frac{ \hat{v}_{\hat{n}j}}{\theta_{\hat{n}}(t)}
   \end{pmatrix}\right.\left.-\1(\Bar{v}_j\neq 0)\E_{\qrob}\left[\prod_{i=1}^{\Hat{n}}\binom{\Hat{X}_i(t)}{\Hat{S}_{ij}}
    \begin{pmatrix} \frac{\hat{X}_1(t)}{\theta_1(t)} - \frac{\hat{X}_{\hat{n}}(t)}{\theta_{\hat{n}}(t)}\\
   \vdots\\
   \frac{\hat{X}_{\hat{n}-1}(t)}{\theta_{\hat{n}-1}(t)}-\frac{\hat{X}_{\hat{n}}(t)}{\theta_{\hat{n}}(t)}
   \end{pmatrix}\right]\right).
\end{align*}

The expectations for the multinomial distribution  are given by
\begin{align*}
    \E_{\qrob}\left[\prod_{i=1}^{\Hat{n}}\binom{\Hat{X}_i}{\Hat{S}_{ij}}\right] = \prod_{i=1}^{\hat{n}}\left(\frac{\theta_i^{\hat{S}_{ij}}}{\hat{S}_{ij}!}\right)\frac{N!}{(N-\sum_{i=1}^{\hat{n}}\hat{S}_{ij})!},
\end{align*}
\begin{align*}
    \E_{\qrob}\left[\prod_{i=1}^{\Hat{n}}\binom{\Hat{X}_i}{\Hat{S}_{ij}}\hat{X}_l\right]= \prod_{i=1}^{\hat{n}}\left(\frac{\theta_i^{\hat{S}_{ij}}}{\hat{S}_{ij}!}\right)\frac{N!}{(N-\sum_{i=1}^{\hat{n}}\hat{S}_{ij})!}\left(\hat{S}_{lj}+\theta_l(N-\sum_{i=1}^{\hat{n}}\hat{S}_{ij})\right).
\end{align*}

Inserting these expectations, we get the following for the drift:
\begin{align*}
       \frac{\mathrm d }{\mathrm d t}\theta(t) = &F(\theta(t))^{-1}\sum_{j=1}^Rc_j\prod_{i=1}^{\Bar{n}}\binom{y_i}{\Bar{S}_{ij}} \prod_{i=1}^{\hat{n}}\left(\frac{\theta_i^{\hat{S}_{ij}}(t)}{\hat{S}_{ij}!}\right)\frac{N!}{(N-\sum_{i=1}^{\hat{n}}\hat{S}_{ij})!}\\
       &\left(\1(\Bar{v}_j=0)\begin{pmatrix} \frac{\hat{v}_{1j}}{\theta_1(t)} - \frac{\hat{v}_{\hat{n}j}}{\theta_{\hat{n}}(t)}\\
   \vdots\\
   \frac{\hat{v}_{\hat{n}-1,j}}{\theta_{\hat{n}-1}(t)}- \frac{ \hat{v}_{\hat{n}j}}{\theta_{\hat{n}}(t)}
   \end{pmatrix}-\1(\Bar{v}_j\neq 0)
    \begin{pmatrix} \frac{\hat{S}_{1j}}{\theta_1(t)} - \frac{\hat{S}_{\hat{n}j} }{\theta_{\hat{n}}(t)}\\
   \vdots\\
   \frac{\hat{S}_{\hat{n}-1,j}}{\theta_{\hat{n}-1}(t)}-\frac{\hat{S}_{\hat{n}j} }{\theta_{\hat{n}}(t)}
   \end{pmatrix}\right).
\end{align*}

The fisher matrix and its inverse are given by
\begin{equation*}
    F(\theta)_{ij} = \frac{\hat{N}}{\theta_{\hat{n}}} + \delta_{ij}\frac{\hat{N}}{\theta_i}
\end{equation*}\begin{equation*}
    F(\theta)^{-1}_{ij} = \frac{1}{\hat{N}^2}\text{Cov}_{ij},
\end{equation*}
where $\text{Cov}_{ij}$ are the elements of the covariance matrix of the multinomial distribution for $1\leq i,j \leq {\hat{n}}-1$.
This leads to the drift:

\begin{align*}
    \frac{\mathrm d \theta_l(t)}{\mathrm d t}  &= \sum_{j=1}^Rc_j\prod_{i=1}^{\Bar{n}}\binom{y_i}{\Bar{S}_{ij}} \prod_{i=1}^{\hat{n}}\left(\frac{\theta_i^{\hat{S}_{ij}}(t)}{\hat{S}_{ij}!}\right)\frac{N!}{(N-\sum_{i=1}^{\hat{n}}\hat{S}_{ij})!}\\
    &\qquad\left(\sum_{i=1}^{\hat{n}-1}\frac{1}{N}(-\theta_l(t)\theta_i(t))\left(\1(\Bar{v}_j=0)(\frac{\hat{v}_{ij}}{\theta_i(t)}-\frac{\hat{v}_{\hat{n}j}}{\theta_{\hat{n}}(t)})
    -\1(\Bar{v}_j\neq 0)(\frac{\hat{S}_{ij}}{\theta_i(t)}-\frac{\hat{S}_{\hat{n}j}}{\theta_{\hat{n}}(t)})
    \right)\right.\\ 
   &\qquad\left. +\frac{1}{N}(\theta_l(t))\left(\1(\Bar{v}_j=0)(\frac{\hat{v}_{lj}}{\theta_l(t)}-\frac{\hat{v}_{\hat{n}j}}{\theta_{\hat{n}}(t)})
    -\1(\Bar{v}_j\neq 0)(\frac{\hat{S}_{lj}}{\theta_l(t)}-\frac{\hat{S}_{\hat{n}j}}{\theta_{\hat{n}}(t)})
    \right)\right)\\
&= \sum_{j=1}^Rc_j\prod_{i=1}^{\Bar{n}}\binom{y_i}{\Bar{S}_{ij}} \prod_{i=1}^{\hat{n}}\left(\frac{\theta_i^{\hat{S}_{ij}}(t)}{\hat{S}_{ij}!}\right)\frac{N!}{(N-\sum_{i=1}^{\hat{n}}\hat{S}_{ij})!}\\
&\qquad\left(\1(\Bar{v}_j=0)\left(\sum_{i=1}^{\hat{n}-1}\frac{1}{N}(-\theta_l(t))(\hat{v}_{ij}) +\frac{1}{N}\hat{v}_{lj}+\sum_{i=1}^{\hat{n}-1}(\theta_i(t)-1)\frac{1}{N}(\theta_l(t))(\frac{\hat{v}_{nj}}{\theta_{\hat{n}}(t)})
\right) \right.\\
&\qquad\left. - \1(\Bar{v_j} \neq 0)\left(\sum_{i=1}^{\hat{n}-1}\frac{1}{N}(-\theta_l(t))(\hat{S}_{ij}) +\frac{1}{N}\hat{S}_{lj}+\sum_{i=1}^{\hat{n}-1}(\theta_i(t)-1)\frac{1}{N}(\theta_l(t))(\frac{\hat{S}_{nj}}{\theta_{\hat{n}}(t)})
\right) 
\right)\\
&= \sum_{j=1}^Rc_j\prod_{i=1}^{\Bar{n}}\binom{y_i}{\Bar{S}_{ij}} \prod_{i=1}^{\hat{n}}\left(\frac{\theta_i^{\hat{S}_{ij}}(t)}{\hat{S}_{ij}!}\right)\frac{(N-1)!}{(N-\sum_{i=1}^{\hat{n}}\hat{S}_{ij})!}\\
&\qquad\left(\1(\Bar{v}_j=0)\left(\sum_{i=1}^{\hat{n}}(-\theta_l(t))(\hat{v}_{ij}) +\hat{v}_{lj}
\right)  - \1(\Bar{v}_j \neq 0)\left(\sum_{i=1}^{\hat{n}}(-\theta_l(t))(\hat{S}_{ij}) +\hat{S}_{lj}
\right) 
\right).
\end{align*}

For the closed-loop \ac{crn} problem considered in the experiments, we get the following drift:
\begin{equation*}
\begin{aligned}
\frac{\mathrm d \theta_1(t)}{\mathrm d t}=-c_{12}\1(u(t)=0)\theta_1(t)+c_{21}\1(u(t)=1)\theta_2(t) + c_{24}\theta_1(t)\theta_2(t) + c_{13}\theta_1(t)\theta_1(t) - c_{13}\theta_1(t).
\end{aligned}
\end{equation*}
We only need to compute the drift for $\theta_1$ as the parameters always need to satisfiy $\theta_1 +\theta_2 =1$

The filter distribution updates, whenever the observed states jump.
If the change vector $\Bar{v}$ is seen, the exact filtering distribution is given by:

\begin{equation*}
    \pi_{t+}(x) = \frac{\sum_{j=1}^R \1(\Bar{v}=\Bar{v}_j)h_j(\hat{x}-\hat{v}_j,\Bar{x}(t-))\pi_{t-}(\hat{x}-\hat{v}_j,\Bar{x}(t-))}{\sum_{j=1}^R \1(\Bar{v}=\Bar{v}_j)\Eof*{h_j(\hat{X},\Bar{X}(t-))}},
\end{equation*}
where the expectation in the denominator is \wrt the filter distribution before the jump.
We want to approximate the posterior using moment matching. Since we work with the multinomial distribution, we only need to match the first moment $M$. Here we give the equation for the $l$-th element of the first moment after the jump $M_{l,t+}$:

\begin{equation*}
\begin{aligned}
        M_{l,t+} &= \frac{\sum_x \hat{x}_l\sum_{j=1}^R \1(\Bar{v}=\Bar{v}_j)h_j(\hat{x}-\hat{v}_j,\Bar{x}(t-))\pi_{t-}(\hat{x}-\hat{v}_j,\Bar{x}(t-))}{\sum_{j=1}^R \1(\Bar{v}=\Bar{v}_j)\Eof*{h_j(\hat{X},\Bar{X}(t-))}} = \\
        &\qquad\frac{\sum_{j=1}^R \1(\Bar{v}=\Bar{v}_j)\Eof*{(\hat{X}_l+\hat{v}_{lj})h_j(\hat{X},\Bar{X}(t-))}}{\sum_{j=1}^R \1(\Bar{v}=\Bar{v}_j)\Eof*{h_j(\hat{X},\Bar{X}(t-))}}.
\end{aligned}
\end{equation*}
We assume that the filter distribution before the jump belongs to the multinomial family. The expectations are given by:
 \begin{align*}
    \Eof*{h_j(\hat{X},\Bar{X}(t-))} =c_j\prod_{i=1}^{\Bar{n}}\binom{y_i(t-)}{\Bar{S}_{ij}} \prod_{i=1}^{\Hat{n}}\left(\frac{\theta_i^{\hat{S}_{ij}}}{\hat{S}_{ij}!}\right) \frac{N!}{(N-\sum_{i=1}^{\hat{n}}\hat{S}_{ij})!},
\end{align*}

\begin{align*}
    \Eof*{(\hat{X}_l+\hat{v}_{lj})h_j(\hat{X},\Bar{X}(t-))} &=   c_j\prod_{i=1}^{\Bar{n}}\binom{y_i(t-)}{\Bar{S}_{ij}}\prod_{i=1}^{\Hat{n}}\left(\frac{\theta_i^{\hat{S}_{ij}}}{\hat{S}_{ij}!}\right)\\
    &\qquad\frac{N!}{(N-\sum_{i=1}^{\hat{n}}\hat{S}_{ij})!} \left(\hat{S}_{lj}+\hat{v}_{lj}+\theta_l(N-\sum_{i=1}^{\hat{n}}\hat{S}_{ij})\right).
\end{align*}
With this we can compute both the drift and the jump updates in closed-form.
\newpage
\section{\uppercase{Dynamic Programming}}
\label{app:control}

\subsection{Dynamic Programming for Continuous-Time MDPs with Discrete State and Action Spaces}
\label{app:mdp_control}
We define the value function of the underlying \ac{mdp} for a state $x \in \mathcal X$ as the expected cumulative reward under the optimal control, \ie,
\begin{equation*}
    V(x) \coloneqq \max_{u_{[t,\infty)}} \Eof*{\int_t^\infty \frac{1}{\tau} e^{-\frac{s-t}{\tau}} R(X(s),u(s))\diff s | X(t) = x},  
\end{equation*}
where we assume that the admissible control $u(t)$ can depend on the state $X(t)$.

By applying the principle of optimality, we can rewrite the value function as:
\begin{equation*}
\begin{aligned}
  V(x) &= \max_{u_{[t,\infty)}} \Eof*{\int_t^{t+h} \frac{1}{\tau} e^{-\frac{s-t}{\tau}} R(X(s),u(s))\diff s  + \int_{t+h}^{\infty} \frac{1}{\tau} e^{-\frac{s-t}{\tau}} R(X(s),u(s))\diff s | X(t) = x}\\
  &= \max_{u_{[t,t+h)}} \Eof*{\int_t^{t+h} \frac{1}{\tau} e^{-\frac{s-t}{\tau}} R(X(s),u(s))\diff s  + e^{-\frac{h}{\tau}}V(x(t+h))| X(t) = x}.
  \end{aligned}
\end{equation*}

The dynamics of the \ac{ctmc} are defined through the rate function $\Lambda(x,x',u,t)$:
\begin{equation*}
        \Prob(X(t+h)=x' \mid X(t)=x, u(t)=u)=\begin{cases}
            \Lambda(x,x', u, t)h + o(h) & \text{if $x'\neq x$}\\
            1 -\sum_{x'\neq x}\left[\Lambda(x,x', u, t)h + o(h)\right] & \text{if $x'=x$}\\
        \end{cases}.
\end{equation*}

With these we can compute the expectation of $V(x(t+h))$ and therefore reformulate $V(x)$:
\begin{equation*}
\begin{aligned}
    V(x) &= \max_{u_{[t,t+h)}} \Eof*{\int_t^{t+h} \frac{1}{\tau} e^{-\frac{s-t}{\tau}} R(X(s),u(s))\diff s | X(t) = x}\\
    &+ e^{-\frac{h}{\tau}}\left(V(x) + \sum_{x'\neq x}\left[\Lambda(x,x', u, t)h + o(h)\right](V(x')-V(x)) \right).
\end{aligned}
\end{equation*}

By bringing the $V(x)$ terms to the left hand side and dividing by $h$ we get
\begin{equation*}
    \begin{aligned}
    V(x)(\frac{1-e^{-\frac{h}{\tau}}}{h}) &= \max_{u_{[t,t+h)}} \frac{1}{h}\Eof*{\int_t^{t+h} \frac{1}{\tau} e^{-\frac{s-t}{\tau}} R(X(s),u(s))\diff s | X(t) = x}\\
    &+ e^{-\frac{h}{\tau}}\left(\sum_{x'\neq x}\left[\Lambda(x,x', u, t) + \frac{o(h)}{h}\right](V(x')-V(x)) \right).     
    \end{aligned}
\end{equation*}
Taking the limit $\lim_{h \to 0}$ we find the optimality conditions as
\begin{equation*}
    \frac{1}{\tau}V(x) = \max_{u}\frac{1}{\tau}R(x,u) + \sum_{x'\neq x}\Lambda(x,x', u, t)(V(x')-V(x)),
\end{equation*}
where we define the state action value function as
\begin{equation*}
    Q(x,u) = R(x,u) + \tau\sum_{x'\neq x}\Lambda(x,x', u, t)(V(x')-V(x)).
\end{equation*}
In the following steps, we show how we can reformulate the value function as a contraction mapping.

\begin{equation*}
    V(x) = \max_{u}R(x,u) +\tau \sum_{x'\neq x}\Lambda(x,x', u, t)(V(x')-V(x)).
\end{equation*}

\begin{equation*}
    V(x)\left(1 + \tau\sum_{x'\neq x}\Lambda(x,x', u, t)\right) = \max_{u}R(x,u) +\tau \sum_{x'\neq x}\Lambda(x,x', u, t)V(x').
\end{equation*}

\begin{equation*}
    V(x) = \max_{u} \frac{R(x,u)}{1 + \tau\sum_{x'\neq x}\Lambda(x,x', u, t)} +\tau \sum_{x'\neq x}\frac{\Lambda(x,x', u, t)}{1 + \tau\sum_{x'\neq x}\Lambda(x,x', u, t)}V(x').
\end{equation*}

With $u^\ast$ as the maximizer of the right hand side, we can also write it in the following form:

\begin{equation*}
    Q(x,u^\ast) =\frac{R(x,u^\ast)}{1 + \tau\sum_{x'\neq x}\Lambda(x,x', u^\ast, t)} +\tau \sum_{x'\neq x}\frac{\Lambda(x,x', u^\ast, t)}{1 + \tau\sum_{x'\neq x}\Lambda(x,x', u^\ast, t)}\max_{u'}Q(x',u').
\end{equation*}

\subsection{Q-Learning}
\label{app:Q_Learn}
We built our approximation of the state action value function on the Q-Learning method by \citet{bradtke1994reinforcement}. In their work they extended traditional reinforcement learning methods, originally designed for discrete-time \acp{mdp}, to encompass \acp{smdp}. In \acp{smdp} the process dynamics are described by semi-Markov processes, which are a generalization of \acs{ctmc}. Equivalently, our methodology is easily adaptable to scenarios, where the fully observed problem can be described by a \ac{smdp}.

In their method, transitions from state $x$ to state $x'$ are sampled while selecting action $u$. Subsequently, the state-action value function $Q(x,u)$ is updated using the information from the sampled transition and the associated reward $R(x,u)$.
In contrast, in our method, we forego the sampling of transitions and, instead, aggregate over all possible transitions, harnessing complete knowledge of the \ac{mdp}. This is possible, since the possible transitions for each state-action pair are finite.

We use a fully connected neural network $Q(x,u;\theta)$ to approximate the state-action value function.
To enhance stability in the learning process, the utilization of a target network $\hat{Q}(x,u;\theta^-)$ is a viable strategy. This secondary network is updated at a slower rate compared to the original network.
\cref{alg:q_learning} shows the pseudo-code for this Q-Learning method.

\begin{algorithm}
\caption{Q-Learning without Transition Sampling}
\label{alg:q_learning}
\begin{algorithmic}
\State Initialize state-action value function $Q$ with random weights $\theta$
\State Initialize target state-action value function $\hat{Q}$ with weights $\theta^-=\theta$
\State Set target update parameter $\kappa$
\For{episode = 1, M}
    \State Sample a batch of states $x$
    \State Get rates $\Lambda(x,x',u,t)$ for all possible next states $x'$ and all actions $u$ 
    \State Compute the target $y(x,u) = R(x,u) + \tau\sum_{x'\neq x}\Lambda(x,x', u, t)(\max_{u'}\hat{Q}(x',u';\theta^-)-\max_{u'}\hat{Q}(x,u';\theta^-))$
    \State Perform a gradient step on the mean squared error between $Q(x, u; \theta)$ and $y$ with respect to $\theta$
    \State Update the target network: $\theta' \leftarrow \kappa \theta + (1 - \kappa) \theta'$
\EndFor
\end{algorithmic}
\end{algorithm}
\newpage
\section{\uppercase{Experiments}}
\label{app:experiments}
\subsection{Additional Information on Experiments}
\begin{figure}[h]
    \centering
    \includegraphics{figures/AISTATS_figures/LV/Lgend_LV_aistats.pdf}
        \includegraphics{figures/AISTATS_figures/LV/Lgend_LV_aistats.pdf}
    \includegraphics{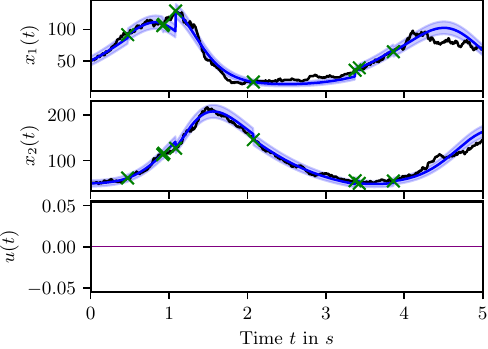}
    \includegraphics{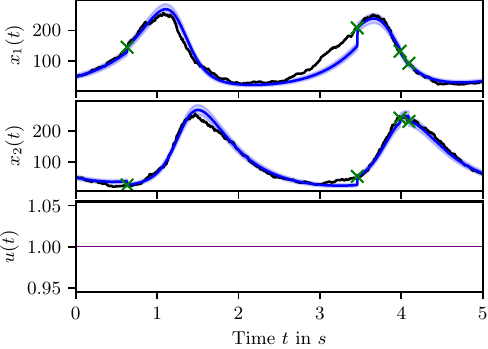}
    \caption{Sample trajectory for the \ac{LV} problem with constant control.}
    \label{fig:add_LV}
\end{figure}

The reward function for the considered experiments in the main section is given by $R(x,u)=-\frac{1}{n}\sum_{i=1}^n(\frac{x_i-x^\ast_i}{l})^2$. 
For the problems with discrete-time measurements we chose Gaussian noise measurements of the exact states, $\mathcal{N}(y\mid x, \sigma^2 I)$. For the queueing problem we observe measurements of queue $2$ and queue $3$, while the first queue is unobserved. For the \ac{LV} problem both species are observed.
The parameters for the experiments are given in the tables below.
In \cref{fig:add_LV} we provide sample trajectories for the \ac{LV} problem with a constant control. By comparing these with the results in the main section, we can see that the our control method effectively combines the dynamics of both actions,
leading to trajectories that are closer to the goal state.
\begin{table}[ht]
    \centering
    \caption{Parameter of the queueing problem}
    \begin{tabular}{|c|c|}
        \hline
        \textbf{Parameter} & \textbf{Value} \\
        \hline
        number of queues $n$ & 3 \\
        buffer size $N$ & 1000 \\
        arrival rate $\lambda_1$ & 10.0 \\
        arrival rate $\lambda_2$ & 10.0 \\
        service rate $\mu_1$ & 20.0 \\
        service rate $\mu_2$ & 20.0 \\
        service rate $\mu_3$ & 20.0 \\
        reward scale $l$ & 100.0\\
        goal state  $x^\ast$ & $[0, 0, 0]^\top$\\
        discount $\tau$ & 5.0\\
        observation noise $\sigma^2$ & 5.0\\
        \hline
    \end{tabular}
\end{table}

\begin{table}[ht]
    \centering
    \caption{Parameter of the \ac{LV} problem}
    \begin{tabular}{|c|c|}
        \hline
        \textbf{Parameter} & \textbf{Value} \\
        \hline
        number of species $n$ & 2 \\
        $c_1$  & 2.5 \\
        $c_2$ & 0.025 \\
        $c_3(u=0)$  & 1.25\\
        $c_3(u=1)$ & 2.5 \\
        reward scale $l$ & 20.0\\
        goal state $x^{\ast}$ & $[100, 100]^T$\\
        discount $\tau$ & 5.0\\
        observation noise $\sigma^2$ & 5.0\\
        \hline
    \end{tabular}
\end{table}

\begin{table}[ht]
    \centering
    \caption{Parameter of the closed-loop \ac{crn} problem}
    \begin{tabular}{|c|c|}
        \hline
        \textbf{Parameter} & \textbf{Value} \\
        \hline
        number of species $n$ & 4 \\
        total species number $N$ & 300 \\
        $c_{12}$  & 0.05 \\
       $c_{21}$  & 0.05 \\
        $c_{13}$  & 0.05 \\
       $c_{31}$  & 0.05 \\
        $c_{24}$  & 0.05 \\
       $c_{42}$  & 0.05 \\
        $c_{34}$  & 0.05 \\
       $c_{43}$  & 0.05 \\
       goal state  $x^\ast$ & $[75, 75, 75, 75]^\top$\\
       reward scale $l$ & 1.0\\
         discount     $\tau$  & 5.0 \\
        \hline
    \end{tabular}
\end{table}

\newpage
\subsection{Projection Filter vs Exact Filter}
\begin{figure}
\centering
\begin{minipage}{0.5\textwidth}   
\includegraphics{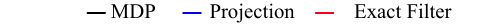}
\includegraphics{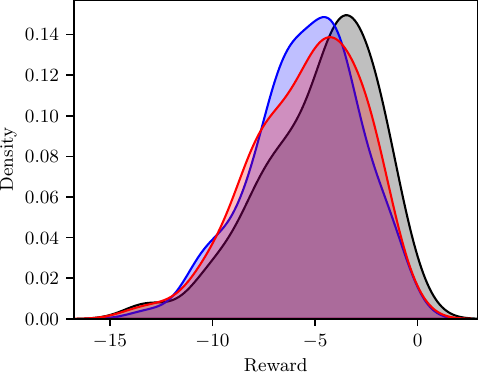}
    \caption{Kernel density estimates of the cumulative reward for different policies using 100 samples.}
    \label{fig:comparison}
    \end{minipage}
\end{figure}

To be able to evaluate the effect the projection filter has on the control method, we compare it to a policy which employs the QMDP method with respect to the exact filtering distribution.
Because exact filtering is only tractable on small state spaces, we create a simpler queueing example based on the experiment in the main section with parameters given in \cref{paramQ}.
For this problem we run $100$ sample trajectories for each of the following methods:
\begin{itemize}
\item the QMDP method based on the projection filter,
\item the QMDP method based on the exact filter,
\item and an optimal controller with full knowledge of the state.
\end{itemize}
For the projection filter we choose again the product binomial distribution as described in \cref{sec:experiments}.
\cref{fig:comparison} shows the kernel density estimates of the cumulative reward for the different polices based on the sample trajectories. Overall the results of all policies are very similar, likely due to the relatively modest scale of the problem at hand. Still we can see that the optimal controller with full knowledge performs best, which is attributable to the fact that the other two methods only have partial observations of the system.
The method based on the exact filtering only performs slightly better than the method based on the projection filter. This shows, that the performance loss of using the projection method is very small, making it a reasonable choice for larger problems, where the exact filtering method is intractable.
\begin{table}[ht]
\centering
    \caption{Parameter of the queueing problem}
    \begin{tabular}{|c|c|}
        \hline
        \textbf{Parameter} & \textbf{Value} \\
        \hline
        number of queues $n$ & 3 \\
        buffer size $N$ & 5 \\
        arrival rate $\lambda_1$ & 1.0 \\
        arrival rate $\lambda_2$ & 1.0 \\
        service rate $\mu_1$ & 2.0 \\
        service rate $\mu_2$ & 2.0 \\
        service rate $\mu_3$ & 2.0 \\
        reward scale $l$ & 1.0\\
        goal state  $x^\ast$ & $[0, 0, 0]^\top$\\
        discount $\tau$ & 5.0\\
        observation noise $\sigma^2$ & 0.5\\
        \hline
    \end{tabular}
    \label{paramQ}
\end{table}

\end{document}